# Optimal Value of Information in Graphical Models

**Andreas Krause**        KRAUSEA@CALTECH.EDU
*California Institute of Technology,*
*1200 E California Blvd.,*
*Pasadena, CA 91125 USA*

**Carlos Guestrin**        GUESTRIN@CS.CMU.EDU
*Carnegie Mellon University,*
*5000 Forbes Ave.,*
*Pittsburgh, PA 15213 USA*

## Abstract

Many real-world decision making tasks require us to choose among several expensive observations. In a sensor network, for example, it is important to select the subset of sensors that is expected to provide the strongest reduction in uncertainty. In medical decision making tasks, one needs to select which tests to administer before deciding on the most effective treatment. It has been general practice to use heuristic-guided procedures for selecting observations. In this paper, we present the first efficient optimal algorithms for selecting observations for a class of probabilistic graphical models. For example, our algorithms allow to optimally label hidden variables in Hidden Markov Models (HMMs). We provide results for both selecting the optimal subset of observations, and for obtaining an optimal conditional observation plan.

Furthermore we prove a surprising result: In most graphical models tasks, if one designs an efficient algorithm for chain graphs, such as HMMs, this procedure can be generalized to polytree graphical models. We prove that the optimizing value of information is $\mathbf{NP^{PP}}$-hard even for polytrees. It also follows from our results that just computing decision theoretic value of information objective functions, which are commonly used in practice, is a $\#\mathbf{P}$-complete problem even on Naive Bayes models (a simple special case of polytrees).

In addition, we consider several extensions, such as using our algorithms for scheduling observation selection for multiple sensors. We demonstrate the effectiveness of our approach on several real-world datasets, including a prototype sensor network deployment for energy conservation in buildings.

## 1. Introduction

In probabilistic reasoning, where one can choose among several possible but expensive observations, it is often a central issue to decide which variables to observe in order to most effectively increase the expected utility (Howard, 1966; Howard & Matheson, 1984; Mookerjee & Mannino, 1997; Lindley, 1956). In a medical expert system, for example, multiple tests are available, and each test has a different cost (Turney, 1995; Heckerman, Horvitz, & Middleton, 1993). In such systems, it is thus important to decide which tests to perform in order to become most certain about the patient's condition, at a minimum cost. Occasionally, the cost of testing can even exceed the value of information for any possible outcome, suggesting to discontinue any further testing.

The following running example motivates our research and is empirically evaluated in Section 6. Consider a temperature monitoring task, where wireless temperature sensors are distributed across a building. The task is to become most certain about the temperature distribution, whilst minimizing energy expenditure, a critically constrained resource (Deshpande, Guestrin, Madden, Hellerstein, & Hong, 2004). Such fine-grained building monitoring is required to obtain significant energy savings (Singhvi, Krause, Guestrin, Garrett, & Matthews, 2005).

Many researchers have suggested the use of myopic (greedy) approaches to select observations (Scheffer, Decomain, & Wrobel, 2001; van der Gaag & Wessels, 1993; Dittmer & Jensen, 1997; Bayer-Zubek, 2004; Kapoor, Horvitz, & Basu, 2007). Unfortunately, in general, this heuristic does not provide any performance guarantees. In this paper, we present efficient algorithms, which guarantee optimal *nonmyopic* value of information in *chain graphical models*. For example, our algorithms can be used for optimal active labeling of hidden states in Hidden Markov Models (HMMs, Baum & Petrie, 1966). We address two settings: *subset selection*, where the optimal subset of observations is obtained in an open-loop fashion, and *conditional plans*, a sequential, closed-loop plan where the observation strategy depends on the actual value of the observed variables (*c.f.,* Figure 1). To our knowledge, these are the first optimal and efficient algorithms for observation selection and diagnostic planning based on value of information for this class of graphical models. For both settings, we address the *filtering* and the *smoothing* versions: Filtering is important in online decision making, where our decisions can only utilize observations made in the past. Smoothing arises for example in structured classification tasks, where there is no temporal dimension in the data, and hence all observations can be taken into account. We call our approach VoIDP as the algorithms use Dynamic Programming to optimize Value of Information. We evaluate our VoIDP algorithms empirically on three real-world datasets, and also show that they are well-suited for interactive classification of sequential data.

Most inference problems in graphical models, such as computing marginal distributions and finding the most probable explanation, that can be solved efficiently for chain-structured graphs, can also be solved efficiently for polytrees. We prove that the problem of selecting the best $k$ observations for maximizing decision theoretic value of information is $\mathbf{NP^{PP}}$-complete even for discrete polytree graphical models, giving a complexity theoretic classification of a core artificial intelligence problem. $\mathbf{NP^{PP}}$-complete problems are believed to be significantly harder than $\mathbf{NP}$-complete or even #$\mathbf{P}$-complete problems commonly arising in the context of graphical models. We furthermore prove that just evaluating decision-theoretic value of information objective functions is #$\mathbf{P}$-complete even in the case of Naive Bayes models, a simple special case of polytree graphical models that is frequently used in practice (*c.f.,* Domingos & Pazzani, 1997).

Unfortunately, these hardness results show that, while the problem of scheduling a single sensor can be optimally solved using our algorithms, the problem of scheduling multiple, correlated sensors is wildly intractable. Nevertheless, we show how our VoIDP algorithms for single sensor scheduling can be used to approximately optimize a multi-sensor schedule. We demonstrate the effectiveness of this approach on a real sensor network testbed for building management.

In summary, we provide the following contributions:

- We present the first optimal algorithms for nonmyopically computing and optimizing value of information on chain graphical models.

- We show that optimizing decision theoretic value of information is $\mathbf{NP^{PP}}$-hard for discrete polytree graphical models. Just computing decision theoretic value of information is #$\mathbf{P}$-hard even for Naive Bayes models.

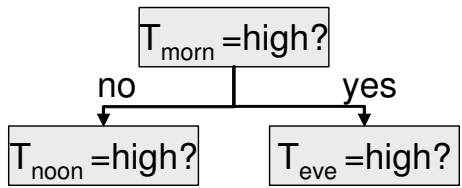

Figure 1: Example of a conditional plan.

- We present several extensions of our algorithms, e.g., to tree graphical models with few leaves, and to multiple correlated chains (for multi-sensor scheduling).

- We extensively evaluate our algorithms on several real-world problems, including sensor scheduling on a real sensor testbed and active labeling in bioinformatics and Natural Language Processing.

## 2. Problem Statement

We will assume that the state of the world is described by a collection of random variables $\mathcal{X}_{\mathcal{V}} = (\mathcal{X}_1, \ldots, \mathcal{X}_n)$, where $\mathcal{V}$ is an index set. For example, $\mathcal{V}$ could denote a set of locations, and $\mathcal{X}_i$ models the temperature reading of a sensor placed at location $i \in \mathcal{V}$. For a subset $\mathcal{A} = \{i_1, \ldots, i_k\} \subseteq \mathcal{V}$, we use the notation $\mathcal{X}_{\mathcal{A}}$ to refer to the random vector $\mathcal{X}_{\mathcal{A}} = (\mathcal{X}_{i_1}, \ldots, \mathcal{X}_{i_k})$. While some of our algorithms extend to continuous distributions, we generally assume that the variables $\mathcal{X}_{\mathcal{V}}$ are discrete. We take a Bayesian approach, and assume a prior probability distribution $P(\mathcal{X}_{\mathcal{V}})$ over the outcomes of the variables. Suppose we select a subset of the variables, $\mathcal{X}_{\mathcal{A}}$ (for $\mathcal{A} \subseteq \mathcal{V}$), and observe $\mathcal{X}_{\mathcal{A}} = \mathbf{x}_{\mathcal{A}}$. For example, $\mathcal{A}$ is the set of locations where we place sensors, or a set of medical tests we decide to perform. After observing the realization of these variables $\mathcal{X}_{\mathcal{A}} = \mathbf{x}_{\mathcal{A}}$, we can compute the posterior distribution over all variables $P(\mathcal{X}_{\mathcal{V}} \mid \mathcal{X}_{\mathcal{A}} = \mathbf{x}_{\mathcal{A}})$. Based on this posterior probability we obtain a reward $R(P(\mathcal{X}_{\mathcal{V}} \mid \mathcal{X}_{\mathcal{A}} = \mathbf{x}_{\mathcal{A}}))$. For example, this reward function could depend on the uncertainty (e.g., measured by the entropy) of the distribution $P(\mathcal{X}_{\mathcal{V}} \mid \mathcal{X}_{\mathcal{A}} = \mathbf{x}_{\mathcal{A}})$. We will describe several examples in more detail below.

In general, when selecting observation, we will not know ahead of time what observations we will make. Instead, we only have a distribution over the possible observations. Hence, we will be interested in the *expected reward*, where we take the expectation over the possible observations.

When optimizing the selection of variables, we can consider different settings: In *subset selection*, our goal is to pick a subset $\mathcal{A}^* \subseteq \mathcal{V}$ of the variables, maximizing

$$\mathcal{A}^* = \operatorname*{argmax}_{\mathcal{A}} \sum_{\mathbf{x}_{\mathcal{A}}} P(\mathcal{X}_{\mathcal{A}} = \mathbf{x}_{\mathcal{A}}) R(P(\mathcal{X}_{\mathcal{V}} \mid \mathcal{X}_{\mathcal{A}} = \mathbf{x}_{\mathcal{A}})), \tag{1}$$

where we impose some constraints on the set $\mathcal{A}$ we are allowed to pick (e.g., on the number of variables that can be selected, etc.). In the subset selection setting, we commit to the selection of the variables before we get to see their realization.

Instead, we can also *sequentially* select one variable after the other, letting our choice depend on the observations made in the past. In this setting, we would like to find a *conditional plan* $\pi^*$

that maximizes

$$\pi^* = \operatorname*{argmax}_{\pi} \sum_{\mathbf{x}_{\mathcal{V}}} P(\mathbf{x}_{\mathcal{V}}) R(P(\mathcal{X}_{\mathcal{V}} \mid \mathcal{X}_{\pi(\mathbf{x}_{\mathcal{V}})} = \mathbf{x}_{\pi(\mathbf{x}_{\mathcal{V}})})). \tag{2}$$

Hereby, $\pi$ is a conditional plan that can select a different set of variables for each possible state of the world $\mathbf{x}_{\mathcal{V}}$. We use the notation $\pi(\mathbf{x}_{\mathcal{V}}) \subseteq \mathcal{V}$ to refer to the subset of variables selected by the conditional plan $\pi$ in state $\mathcal{X}_{\mathcal{V}} = \mathbf{x}_{\mathcal{V}}$. Figure 1 presents an example of a conditional plan for the temperature monitoring example. We will define the notion of conditional planning more formally in Section 4.2.

This general setup of selecting observations goes back in the decision analysis literature to the notion of value of information by Howard (1966) and in the statistical literature to the notion of Bayesian Experimental Design by Lindley (1956). In this paper, we refer to the Problems (1) and (2) as the problems of *optimizing value of information*.

In this paper, we show how the complexity of solving these value of information problems depend on the properties of the probability distribution $P$. We give the first algorithms for optimally solving value of information for an interesting and challenging class of distributions including Hidden Markov Models. We also present hardness results showing that optimizing value of information is wildly intractable ($\mathbf{NP^{PP}}$-complete) even for probability distributions for which efficient inference is possible (even for Naive Bayes models and discrete polytrees).

## 2.1 Optimization Criteria

In this paper, we will consider a class of *local reward*[1] functions $R_i$, which are defined on the marginal probability distributions of the variables $\mathcal{X}_i$. This class has the computational advantage that local rewards can be evaluated using probabilistic inference techniques. The total reward will then be the sum of all local rewards.

Let $\mathcal{A}$ be a subset of $\mathcal{V}$. Then $P(\mathcal{X}_j \mid \mathcal{X}_{\mathcal{A}} = \mathbf{x}_{\mathcal{A}})$ denotes the marginal distribution of variable $\mathcal{X}_j$ conditioned on observations $\mathcal{X}_{\mathcal{A}} = \mathbf{x}_{\mathcal{A}}$. For example, in our temperature monitoring application, $\mathcal{X}_j$ models the temperature at location $j \in \mathcal{V}$. The conditional marginal distribution $P(\mathcal{X}_j = x_j \mid \mathcal{X}_{\mathcal{A}} = \mathbf{x}_{\mathcal{A}})$ then models the conditional distribution of the temperature at location $j$ after observing the temperature at locations $\mathcal{A} \subseteq \mathcal{V}$.

For classification purposes, it can be more appropriate to consider the *max-marginals*

$$P^{max}(\mathcal{X}_j = \mathbf{x}_j \mid \mathcal{X}_{\mathcal{A}} = \mathbf{x}_{\mathcal{A}}) = \max_{\mathbf{x}_{\mathcal{V}}} P(\mathcal{X}_{\mathcal{V}} = \mathbf{x}_{\mathcal{V}}, \mathcal{X}_j = x_j \mid \mathcal{X}_{\mathcal{A}} = \mathbf{x}_{\mathcal{A}}),$$

that is, for $\mathcal{X}_j$ set to value $x_j$, the probability of the most probable assignment $\mathcal{X}_{\mathcal{V}} = \mathbf{x}_{\mathcal{V}}$ to all other random variables (including $\mathcal{X}_j$ for simplicity of notation) conditioned on the observations $\mathcal{X}_{\mathcal{A}} = \mathbf{x}_{\mathcal{A}}$.

The *local reward* $R_j$ is a functional on the probability distribution $P$ or $P^{max}$ over $\mathcal{X}_j$. That is, $R_j$ takes an entire distribution over the variable $\mathcal{X}_j$ and maps it to a reward value. Typically, the reward functions will be chosen such that "certain" or "peaked" distributions obtain higher reward.

To simplify notation, we write

$$R_j(\mathcal{X}_j \mid \mathbf{x}_{\mathcal{A}}) \triangleq R_j(P(X_j \mid \mathcal{X}_{\mathcal{A}} = \mathbf{x}_{\mathcal{A}}))$$

---

1. Local reward functions are also widely used in additively independent utility models, (*c.f.*, Keeney & Raiffa, 1976).

to denote the reward for variable $\mathcal{X}_j$ upon observing $\mathcal{X}_\mathcal{A} = \mathbf{x}_\mathcal{A}$, and

$$R_j(\mathcal{X}_j \mid \mathcal{X}_\mathcal{A}) \triangleq \sum_{\mathbf{x}_\mathcal{A}} P(\mathcal{X}_\mathcal{A} = \mathbf{x}_\mathcal{A}) R_j(\mathcal{X}_j \mid \mathbf{x}_\mathcal{A})$$

to refer to *expected local rewards*, where the expectation is taken over all assignments $\mathbf{x}_\mathcal{A}$ to the observations $\mathcal{A}$. Important local reward functions include:

**Residual entropy.** If we set

$$R_j(\mathcal{X}_j \mid \mathbf{x}_\mathcal{A}) = -H(\mathcal{X}_j \mid \mathbf{x}_\mathcal{A}) = \sum_{x_j} P(x_j \mid \mathbf{x}_\mathcal{A}) \log_2 P(x_j \mid \mathbf{x}_\mathcal{A}),$$

the objective in the optimization problem becomes to minimize the sum of residual entropies. Optimizing this reward function attempts to reduce the uncertainty in predicting the marginals $\mathcal{X}_i$. We choose this reward function in our running example to measure the uncertainty about the temperature distribution.

**Joint entropy.** Instead of minimizing the sum of residual entropies $\sum_i H(\mathcal{X}_i)$, we can also attempt to minimize the joint entropy of the entire distribution,

$$H(\mathcal{X}_\mathcal{V}) = -\sum_{\mathbf{x}_\mathcal{V}} P(\mathbf{x}_\mathcal{V}) \log_2 P(\mathbf{x}_\mathcal{V}).$$

Note, that the joint entropy depends on the full probability distribution $P(\mathcal{X}_\mathcal{V})$, rather than on the marginals $P(\mathcal{X}_i)$, and hence it is not local. Nevertheless, we can exploit the chain rule for the joint entropy $H(\mathcal{X}_\mathcal{B})$ of a set of random variables $\mathcal{B} = \{1, \ldots, m\}$ (*c.f.,* Cover & Thomas, 1991),

$$H(\mathcal{X}_\mathcal{B}) = H(\mathcal{X}_1) + H(\mathcal{X}_2 \mid \mathcal{X}_1) + H(\mathcal{X}_3 \mid \mathcal{X}_1, \mathcal{X}_2) + \cdots + H(\mathcal{X}_m \mid \mathcal{X}_1, \ldots, \mathcal{X}_{m-1}).$$

Hence, if we choose the local reward functions $R_j(\mathcal{X}_j \mid \mathcal{X}_\mathcal{A}) = -H(\mathcal{X}_j \mid \mathcal{X}_1, \ldots, \mathcal{X}_{j-1}, \mathcal{X}_\mathcal{A})$, we can optimize a non-local reward function (the joint entropy) using only local reward functions.

**Decision-theoretic value of information.** The concept of local reward functions also includes the concept of decision theoretic value of information. The notion of value of information is widely used (*c.f.,* Howard, 1966; Lindley, 1956; Heckerman et al., 1993), and is formalized, e.g., in the context of influence diagrams (Howard & Matheson, 1984) and Partially Observable Markov Decision Processes (POMDPs, Smallwood & Sondik, 1973). For each variable $\mathcal{X}_j$, let $\mathcal{A}_j$ be a finite set of actions. Also, let $U_j : A_j \times \mathrm{dom}\, X_j \to \mathbb{R}$ be a utility function mapping an action $a \in \mathcal{A}_j$ and an outcome $x \in \mathrm{dom}\, \mathcal{X}_j$ to a real number. The *maximum expected utility principle* states that actions should be selected as to maximize the expected utility,

$$EU_j(a \mid \mathcal{X}_\mathcal{A} = \mathbf{x}_\mathcal{A}) = \sum_{x_j} P(x_j \mid \mathbf{x}_\mathcal{A}) U_j(a, x_j).$$

The more certain we are about $\mathcal{X}_j$, the more economically we can choose our action. This idea is captured by the notion of value of information, where we choose our local reward function

$$R_j(\mathcal{X}_j \mid \mathbf{x}_\mathcal{A}) = \max_a EU_j(a \mid \mathbf{x}_\mathcal{A}).$$

**Margin for structured prediction.** We can also consider the margin of confidence:

$$R_j(\mathcal{X}_j \mid \mathbf{x}_\mathcal{A}) = P^{max}(x_j^* \mid \mathbf{x}_\mathcal{A}) - P^{max}(\bar{x}_j \mid \mathbf{x}_\mathcal{A}),$$

where

$$x_j^* = \operatorname*{argmax}_{x_j} P^{max}(x_j \mid \mathbf{x}_\mathcal{A}) \text{ and } \bar{x}_j = \operatorname*{argmax}_{x_j \neq x_j^*} P^{max}(x_j \mid \mathbf{x}_\mathcal{A}),$$

which describes the margin between the most likely outcome and the closest runner up. This reward function is very useful for structured classification purposes, as shown in Section 6.

**Weighted mean-squared error.** If the variables are continuous, we might want to minimize the mean squared error in our prediction. We can do this by choosing

$$R_j(\mathcal{X}_j \mid \mathbf{x}_\mathcal{A}) = -w_j \operatorname{Var}(\mathcal{X}_j \mid \mathbf{x}_\mathcal{A}),$$

where

$$\operatorname{Var}(\mathcal{X}_j \mid \mathbf{x}_\mathcal{A}) = \int P(x_j \mid \mathbf{x}_\mathcal{A}) \left[ x_j - \int x_j' P(x_j' \mid \mathbf{x}_\mathcal{A}) dx_j' \right]^2 dx_j$$

is the conditional variance of $\mathcal{X}_j$ given $\mathcal{X}_\mathcal{A} = \mathbf{x}_\mathcal{A}$, and $w_j$ is a weight indicating the importance of variable $\mathcal{X}_j$.

**Monitoring for critical regions (Hotspot sampling).** Suppose we want to use sensors for detecting fire. More generally, we want to detect, for each $j$, whether $\mathcal{X}_j \in \mathfrak{C}_j$, where $\mathfrak{C}_j \subseteq \operatorname{dom} X_j$ is a "critical region" for variable $\mathcal{X}_j$. Then the local reward function

$$R_j(\mathcal{X}_j \mid \mathbf{x}_\mathcal{A}) = P(\mathcal{X}_j \in \mathfrak{C}_j \mid \mathbf{x}_\mathcal{A})$$

favors observations $\mathcal{A}$ that maximize the probability of detecting critical regions.

**Function optimization (Correlated bandits).** Consider a setting where we have a collection of random variables $\mathcal{X}_\mathcal{V}$ taking numerical values in some interval $[-m, m]$, and, after selecting some of the variables, we get the reward $\sum_i x_i$. This setting arises if we want to optimize an unknown (random) function, where evaluating the function is expensive. In this setting, we are encouraged to only evaluate the function where it is likely to obtain high values. We can maximize our expected total reward if we choose the local reward function

$$R_j(\mathcal{X}_j \mid \mathbf{x}_\mathcal{A}) = \int x_j P(x_j \mid \mathbf{x}_\mathcal{A}) dx_j,$$

i.e., the expectation of variable $\mathcal{X}_j$ given observations $\mathbf{x}_\mathcal{A}$. This setting of optimizing a random function can also be considered a version of the classical $k$-armed bandit problem with correlated arms. More details about the relationship with bandit problems are given in Section 8.

These examples demonstrate the generality of our notion of local reward. Note that most examples apply to continuous distributions just as well as for discrete distributions.

## 2.2 Cost of Selecting Observations

We also want to capture the constraint that observations are expensive. This can mean that each observation $\mathcal{X}_j$ has an associated positive *penalty* $C_j$ that effectively decreases the reward. In our example, we might be interested in trading off accuracy with sensing energy expenditure. Alternatively, it is also possible to define a *budget* $B$ for selecting observations, where each one is associated with an integer *cost* $\beta_j$. Here, we want to select observations whose sum cost is within the budget, but these costs do not decrease the reward. In our running example, the sensors could be powered by solar power, and regain a certain amount of energy per day, which allows a certain amount of sensing. Our formulation of the optimization problems allows both for penalties and budgets. To simplify notation we also write $C(\mathcal{A}) = \sum_{j \in \mathcal{A}} C_j$ and $\beta(\mathcal{A}) = \sum_{j \in \mathcal{A}} \beta_j$ to extend $C$ and $\beta$ to sets.

Instead of fixed penalties and costs per observation, both can also depend on the state of the world. For example, in the medical domain, applying a particular diagnostic test can bear different risks for the health of the patient, depending on the patient's illness. The algorithms we will develop below can be adapted to accommodate such dependencies in a straight-forward manner. We will present details only for the conditional planning algorithm in Section 4.2.

## 3. Decomposing Rewards

In this section, we will present the key observation that allows us to develop efficient algorithms for nonmyopically optimizing value of information in the class of chain graphical models. The algorithms will be presented in Section 4.

The set of random variables $\mathcal{X}_{\mathcal{V}} = \{\mathcal{X}_1, \ldots, \mathcal{X}_n\}$ forms a *chain graphical model* (a chain), if $\mathcal{X}_i$ is conditionally independent of $\mathcal{X}_{\mathcal{V} \setminus \{i-1, i, i+1\}}$ given $\mathcal{X}_{i-1}$ and $\mathcal{X}_{i+1}$. Without loss of generality we can assume that the joint distribution is specified by the prior $P(\mathcal{X}_1)$ of variable $\mathcal{X}_1$ and the conditional probability distributions $P(\mathcal{X}_{i+1} \mid \mathcal{X}_i)$. The time series model for the temperature measured by one sensor in our example can be formulated as a chain graphical model. Note that the transition probabilities $P(\mathcal{X}_{i+1} \mid \mathcal{X}_i)$ are allowed to depend on the index $i$ (i.e., the chain models are allowed to be *nonstationary*). Chain graphical models have been extensively used in machine learning and signal processing.

Consider for example a Hidden Markov Model unrolled for $n$ time steps, i.e., $\mathcal{V}$ can be partitioned into the hidden variables $\{\mathcal{X}_1, \ldots, \mathcal{X}_n\}$ and the emission variables $\{\mathcal{Y}_1, \ldots, \mathcal{Y}_n\}$. In HMMs, the $\mathcal{Y}_i$ are always observed and the variables $\mathcal{X}_i$ form a chain. In many applications, some of which are discussed in Section 6, we can observe some of the hidden variables $\mathcal{X}_i$ as well, e.g., by asking an expert, in addition to observing the emission variables. In these cases, the problem of selecting expert labels also belongs to the class of chain graphical models addressed by this paper, since the variables $\mathcal{X}_i$ form a chain conditional on the observed values of the emission variables $\mathcal{Y}_i$. This idea can be generalized to the class of Dynamic Bayesian Networks where the separators between time slices have size one, and only these separators can be selected for observation. This formulation also includes certain conditional random fields (Lafferty, McCallum, & Pereira, 2001) which form chains, conditional on the emission variables (the features).

Chain graphical models originating from time series have additional, specific properties: In a system for online decision making, only observations from the past and present time steps can be taken into account, not observations which will be made in the future. This is generally referred to as the *filtering* problem. In this setting, the notation $P(\mathcal{X}_i \mid \mathcal{X}_{\mathcal{A}})$ will refer to the distribution of $\mathcal{X}_i$ conditional on observations in $\mathcal{X}_{\mathcal{A}}$ prior to and including time $i$. For structured classification

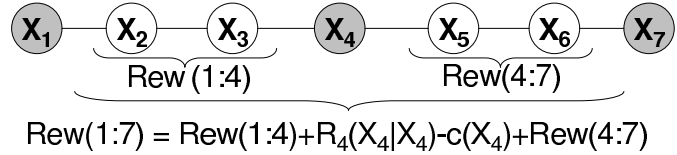

$$\text{Rew}(1{:}7) = \text{Rew}(1{:}4) + R_4(X_4|X_4) - c(X_4) + \text{Rew}(4{:}7)$$

Figure 2: Illustration of the decomposing rewards idea. The reward for chain 1:7 when observing variables $\mathcal{X}_1$, $\mathcal{X}_4$ and $\mathcal{X}_7$ decomposes as the sum of chain 1:4 plus the reward for chain 4:7 plus the immediate reward for observing $\mathcal{X}_4$ minus the cost of observing $\mathcal{X}_4$. Hereby for brevity we use the notation $Rew(a:b) = \sum_{j=a}^{b} R_j(\mathcal{X}_j \mid \mathcal{X}_1, \mathcal{X}_4, \mathcal{X}_7)$.

problems as discussed in Section 6, in general observations made anywhere in the chain must be taken into account. This situation is usually referred to as the *smoothing* problem. We will provide algorithms both for filtering and smoothing.

We will now describe the key insight, which allows for efficient optimization in chains. Consider a set of observations $\mathcal{A} \subseteq \mathcal{V}$. If the $j$ variable is observed, i.e., $j \in \mathcal{A}$, then the local reward is simply $R(\mathcal{X}_j \mid \mathcal{X}_\mathcal{A}) = R(\mathcal{X}_j \mid \mathcal{X}_j)$. Now consider $j \notin \mathcal{A}$, and let $\mathcal{A}_j$ be the subset of $\mathcal{A}$ containing the closest ancestor (and for the smoothing problem also the closest descendent) of $\mathcal{X}_j$ in $\mathcal{X}_\mathcal{A}$. The conditional independence property of the graphical model implies that, given $\mathcal{X}_{\mathcal{A}_j}$, $\mathcal{X}_j$ is independent of the rest of the observed variables, i.e., $P(\mathcal{X}_j \mid \mathcal{X}_\mathcal{A}) = P(\mathcal{X}_j \mid \mathcal{X}_{\mathcal{A}_j})$. Thus, it follows that $R(\mathcal{X}_j \mid \mathcal{X}_\mathcal{A}) = R(\mathcal{X}_j \mid \mathcal{X}_{\mathcal{A}_j})$.

These observations imply that the expected reward of some set of observations decomposes along the chain. For simplicity of notation, we add two independent dummy variables $\mathcal{X}_0$ and $\mathcal{X}_{n+1}$, where $R_0 = C_0 = \beta_0 = R_{n+1} = C_{n+1} = \beta_{n+1} = 0$. Let $\mathcal{A} = \{i_0, \ldots, i_{m+1}\}$ where $i_l < i_{l+1}$, $i_0 = 0$ and $i_{m+1} = n + 1$. Using this notation, the total reward $R(\mathcal{A}) = \sum_j R_j(X_j \mid \mathcal{X}_\mathcal{A})$ for the smoothing case is given by:

$$\sum_{v=0}^{m} \left( R_{i_v}(\mathcal{X}_{i_v} \mid \mathcal{X}_{i_v}) - C_{i_v} + \sum_{j=i_v+1}^{i_{v+1}-1} R_j(\mathcal{X}_j \mid \mathcal{X}_{i_v}, \mathcal{X}_{i_{v+1}}) \right).$$

In filtering settings, we simply replace $R_j(\mathcal{X}_j \mid \mathcal{X}_{i_v}, \mathcal{X}_{i_{v+1}})$ by $R_j(\mathcal{X}_j \mid \mathcal{X}_{i_v})$. Figure 2 illustrates this decomposition.

## 4. Efficient Algorithms for Optimizing Value of Information

In this section, we present algorithms for efficiently and nonmyopically optimizing value of information in chain graphical models.

### 4.1 Efficient Algorithms for Optimal Subset Selection in Chain Models

In the *subset selection* problem, we want to find a most informative subset of the variables to observe *in advance*, i.e., before any observations are made. In our running example, we would, before deploying the sensors, identify $k$ time points that are expected to provide the most informative sensor readings according to our model.

First, define the objective function $L$ on subsets of $\mathcal{V}$ by

$$L(\mathcal{A}) = \sum_{j=1}^{n} R_j(\mathcal{X}_j \mid \mathcal{X}_\mathcal{A}) - C(\mathcal{A}).\tag{3}$$

The *subset selection* problem is to find the optimal subset

$$\mathcal{A}^* = \operatorname*{argmax}_{\mathcal{A} \subseteq \mathcal{V}, \beta(\mathcal{A}) \leq B} L(\mathcal{A})$$

maximizing the sum of expected local rewards minus the penalties, subject to the constraint that the total cost must not exceed the budget $B$.

We solve this optimization problem using a dynamic programming algorithm, where the chain is broken into sub-chains using the insight from Section 3. Consider a sub-chain from variable $\mathcal{X}_a$ to $\mathcal{X}_b$. We define $L_{a:b}^{sm}(k)$ to represent the expected total reward for the sub-chain $\mathcal{X}_a, \ldots, \mathcal{X}_b$, in the smoothing setting where $\mathcal{X}_a$ and $\mathcal{X}_b$ are observed, and with a budget level of $k$. $L_{a:b}^{flt}(k)$ represents the expected reward in the filtering setting where only $\mathcal{X}_a$ is observed. More formally:

$$L_{a:b}^{flt}(k) = \max_{\substack{\mathcal{A} \subset \{a+1\ldots b-1\} \\ \beta(\mathcal{A}) \leq k}} \sum_{j=a+1}^{b-1} R_j(\mathcal{X}_j \mid \mathcal{X}_\mathcal{A}, \mathcal{X}_a) - C(\mathcal{A}),$$

for the filtering version, and

$$L_{a:b}^{sm}(k) = \max_{\substack{\mathcal{A} \subset \{a+1\ldots b-1\} \\ \beta(\mathcal{A}) \leq k}} \sum_{j=a+1}^{b-1} R_j(\mathcal{X}_j \mid \mathcal{X}_\mathcal{A}, \mathcal{X}_a, \mathcal{X}_b) - C(\mathcal{A}),$$

for the smoothing version. Note that in both cases, $L_{0:n+1}(B) = \max_{\mathcal{A}:\beta(\mathcal{A}) \leq B} L(\mathcal{A})$, as in Equation (3), i.e., by computing the values for $L_{a:b}(k)$, we compute the maximum expected total reward for the entire chain.

We can compute $L_{a:b}^{sm}(k)$ and $L_{a:b}^{flt}(k)$ using dynamic programming. The base case is simply:

$$L_{a:b}^{flt}(0) = \sum_{j=a+1}^{b-1} R_j(\mathcal{X}_j \mid \mathcal{X}_a),$$

for filtering, and

$$L_{a:b}^{sm}(0) = \sum_{j=a+1}^{b-1} R_j(\mathcal{X}_j \mid \mathcal{X}_a, \mathcal{X}_b),$$

for smoothing. The recursion for $L_{a:b}(k)$ has two cases: we can choose not to spend any more of the budget, reaching the base case, or we can break the chain into two sub-chains, selecting the optimal observation $\mathcal{X}_j$, where $a < j < b$. In both filtering and smoothing we have

$$L_{a:b}(k) = \max\left\{ L_{a:b}(0), \max_{j:a<j<b, \beta_j \leq k} \{R_j(\mathcal{X}_j \mid \mathcal{X}_j) - C_j + L_{a:j}(0) + L_{j:b}(k - \beta_j)\} \right\}.$$

---

**Input**: Budget $B$, rewards $R_j$, costs $\beta_j$ and penalties $C_j$
**Output**: Optimal selection $\mathcal{A}$ of observation times
**begin**
    **for** $0 \leq a < b \leq n + 1$ **do** compute $L_{a:b}(0)$;
    **for** $k = 1$ **to** $B$ **do**
        **for** $0 \leq a < b \leq n + 1$ **do**
            $sel(-1) := L_{a:b}(0)$;
            **for** $j = a + 1$ **to** $b - 1$ **do** $sel(j) := R_j(\mathcal{X}_j \mid \mathcal{X}_j) - C_j + L_{a:j}(0) + L_{j:b}(k - \beta_j)$;
            $L_{a:b}(k) = \max_{j \in \{a+1,\dots,b-1,-1\}} sel(j)$;
            $\Lambda_{a:b}(k) = \operatorname{argmax}_{j \in \{a+1,\dots,b-1,-1\}} sel(j)$;
        **end**
    **end**
    $a := 0$; $b := n + 1$; $k := B$; $\mathcal{A} := \emptyset$;
    **repeat**
        $j := \Lambda_{a:b}(k)$;
        **if** $j \geq 0$ **then** $\mathcal{A} := \mathcal{A} \cup \{j\}$; $k := k - \beta_j$;
    **until** $j = -1$ ;
**end**

**Algorithm 1**: VoIDP algorithm for optimal subset selection (for both filtering and smoothing).

At first, it may seem that this recursion should consider the optimal split of the budget between the two sub-chains. However, since the subset problem is open-loop and the order of the observations is irrelevant, we only need to consider split points where the first sub-chain receives zero budget.

A pseudo code implementation for this dynamic programming approach, which we call VoIDP *for subset selection* is given in Algorithm 1. The algorithm fills the dynamic programming tables in two loops, the inner loop ranging over all pairs $(a, b)$, $a < b$, and the outer loop increasing $k$. Within the inner loop, when computing the best reward for the sub-chain from $a$ to $b$, it fills out a table $sel$, where $sel(j)$ is the reward that could be obtained by making an observation at $j$, and $sel(-1)$ is the reward if no observation is made.

In addition to computing the optimal rewards $L_{a:b}(k)$ that could be achieved for sub-chain $a : b$ and budget $k$, the algorithm also stores the choices $\Lambda_{a:b}(k)$ that realize this maximum score. Here, $\Lambda_{a:b}(k)$ is the index of the next variable that should be selected for sub-chain $a : b$ with budget $k$, or $-1$ if no variable should be selected. In order to recover the optimal subset for budget $k$, Algorithm 1 uses the quantities $\Lambda_{a:b}$ to recover the optimal subset by tracing the maximal values occurring in the dynamic programming equations. Using an induction proof, we obtain:

**Theorem 1 (Subset Selection).** *The dynamic programming algorithm described above computes the optimal subset with budget $B$ in $(\frac{1}{6}n^3 + \mathcal{O}(n^2))B$ evaluations of expected local rewards.* □

Note that if we do not consider different costs $\beta$ for each variable, we would simply choose $\beta_j = 1$ for all variables and compute $L_{a:b}(N)$. Further note that if the variables $\mathcal{X}_i$ are continuous, our algorithm is still applicable when the integrations and inferences necessary for computing the expected rewards can be performed efficiently. This is the case, for example, in a Gaussian linear model (i.e., the variables $\mathcal{X}_i$ are normally distributed) and the local reward functions are the residual entropies or the residual variances for each variable.

## 4.2 Efficient Algorithms for Optimal Conditional Planning in Chain Models

In the *conditional plan* problem, we want to compute an optimal sequential querying policy $\pi$: We observe a variable, pay the penalty, and depending on all values observed in the past, select the next query, proceeding as long as our budget suffices. The objective is to find the plan with the highest expected reward, where, for each possible sequence of observations, the budget $B$ is not exceeded. For filtering, we can only select observations in the future, whereas in the smoothing case, the next observation can be anywhere in the chain. In our running example, the filtering algorithm would be most appropriate: The sensors would sequentially follow the conditional plan, deciding on the most informative times to sense based on the previous observations. Figure 1 shows an example of such a conditional plan.

### 4.2.1 FROM SUBSET SELECTION TO CONDITIONAL PLANNING

Note that in contrast to the subset selection setting that we considered in Section 4.1, in conditional planning, the set of variables depends on the state of the world $\mathcal{X}_\mathcal{V} = \mathbf{x}_\mathcal{V}$. Hence, for each such state, the conditional plan $\pi$ could select a different set of variables, $\pi(\mathbf{x}_\mathcal{V}) \subseteq \mathcal{V}$. As an example, consider Figure 1, where the set of possible observations is $\mathcal{V} = \{morn, noon, eve\}$, and $\mathcal{X}_\mathcal{V} = \{T_{morn}, T_{noon}, T_{eve}\}$. If the world is in state $\mathbf{x}_\mathcal{V} = (high, low, high)$, then the conditional plan $\pi$ presented in Figure 1 would select $\pi(\mathbf{x}_\mathcal{V}) = \{morn, eve\}$, whereas, if $\mathbf{x}_\mathcal{V} = (low, low, high)$, it would select $\pi(\mathbf{x}_\mathcal{V}) = \{morn, noon\}$. Since the conditional plan is a function of the (random) state of the world, it is a set-valued random variable. In order to optimize Problem (2), we define the objective function[2]

$$\mathrm{J}(\pi) = \sum_{\mathbf{x}_\mathcal{V}} P(\mathbf{x}_\mathcal{V}) \sum_{j=1}^n \left[ R_j(\mathcal{X}_j \mid \mathbf{x}_{\pi(\mathbf{x}_\mathcal{V})}) - C(\pi(\mathbf{x}_\mathcal{V})) \right],$$

i.e., the expected sum of local rewards given the observations made by plan $\pi(\mathbf{x}_\mathcal{V})$ in state $\mathcal{X}_\mathcal{V} = \mathbf{x}_\mathcal{V}$ minus the penalties of the selected variables, where the expectation is taken with respect to the distribution $P(\mathcal{X}_\mathcal{V})$. In addition to defining the value of a policy $\mathrm{J}(\pi)$, we also define the cost $\beta(\pi)$

$$\beta(\pi) = \max_{\mathbf{x}_\mathcal{V}} \beta(\pi(\mathbf{x}_\mathcal{V})),$$

as maximum cost $\beta(\mathcal{A})$ (as defined in Section 2.2) of any set $\mathcal{A} = \pi(\mathbf{x}_\mathcal{V})$ that could be selected by the policy $\pi$, in any state of the world $\mathcal{X}_\mathcal{V} = \mathbf{x}_\mathcal{V}$.

Based on this notation, our goal is to find a policy $\pi^*$ such that

$$\pi^* = \underset{\pi \in \Pi}{\mathrm{argmax}}\, \mathrm{J}(\pi) \text{ such that } \beta(\pi) \leq B,$$

i.e., a policy that has maximum value, and is guaranteed to never have cost exceeding our budget $B$. Hereby $\Pi$ is the class of *sequential* policies (i.e., those, where the observations are chosen sequentially, only based on observations that have been previously made).

It will be useful to introduce the following notation:

$$J(\mathbf{x}_\mathcal{A}; k) = \max_{\pi \in \Pi} \mathrm{J}(\pi \mid \mathcal{X}_\mathcal{A} = \mathbf{x}_\mathcal{A}) \text{ such that } \beta(\pi) \leq k, \tag{4}$$

---

2. Recall that, in the filtering setting, $R(\mathcal{X}_j \mid \mathbf{x}_{\pi(\mathbf{x}_\mathcal{V})}) \triangleq R(\mathcal{X}_j \mid \mathbf{x}_{\mathcal{A}'})$, where $\mathcal{A}' = \{t \in \pi(\mathbf{x}_\mathcal{V}) \text{ s.t. } t \leq j\}$, i.e., only observations from the past are taken into account.

where

$$\mathrm{J}(\pi \mid \mathcal{X}_{\mathcal{A}} = \mathbf{x}_{\mathcal{A}}) = \sum_{\mathbf{x}_{\mathcal{V}}} P(\mathbf{x}_{\mathcal{V}} \mid \mathcal{X}_{\mathcal{A}} = \mathbf{x}_{\mathcal{A}}) \sum_{j=1}^{n} \left[ R_j(\mathcal{X}_j \mid \mathbf{x}_{\pi(\mathbf{x}_{\mathcal{V}})}) - C(\pi(\mathbf{x}_{\mathcal{V}})) \right].$$

Hence, $J(\mathbf{x}_{\mathcal{A}}; k)$ is the best possible reward that can be achieved by any sequential policy with cost at most $k$, after observing $\mathcal{X}_{\mathcal{A}} = \mathbf{x}_{\mathcal{A}}$. Using this notation, our goal is to find the optimal plan with reward $J(\emptyset; B)$.

The value function $J$ satisfies the following recursion. The base case considers the exhausted budget:

$$J(\mathbf{x}_{\mathcal{A}}; 0) = \sum_{j \in \mathcal{V}} R_j(\mathcal{X}_j \mid \mathbf{x}_{\mathcal{A}}) - C(\mathcal{A}).$$

For the recursion, it holds that

$$J(\mathbf{x}_{\mathcal{A}}; k) = \max \left\{ J(\mathbf{x}_{\mathcal{A}}; 0), \max_{j \notin \mathcal{A}} \left\{ \sum_{x_j} P(x_j \mid \mathbf{x}_{\mathcal{A}}) J(\mathbf{x}_{\mathcal{A}}, x_j; k - \beta_j) \right\} \right\}, \tag{5}$$

i.e., the best one can do in state $\mathcal{X}_{\mathcal{A}} = \mathbf{x}_{\mathcal{A}}$ with budget $k$ is to either stop selecting variables, or chose the best next variable and act optimally thereupon.

Note that we can easily allow the cost $\beta_j$ depend on the state $x_j$ of variable $\mathcal{X}_j$. In this case, we would simply replace $\beta_j$ by $\beta_j(x_j)$, and define $J(\mathcal{X}_{\mathcal{A}}, r) = -\infty$ whenever $r < 0$. Equivalently, we can let the penalty $C(\mathcal{A})$ depend on the state by replacing $C(\mathcal{A})$ by $C(\mathbf{x}_{\mathcal{A}})$.

**Relationship to finite-horizon Markov Decision Processes (MDPs).** Note that the function $J(\mathbf{x}_{\mathcal{A}}; k)$ defined in (4) is analogous to the concept of a *value function* in Markov Decision Processes (*c.f.*, Bellman, 1957): In finite-horizon MDPs, the value function $V(s; k)$ models the maximum expected reward obtainable when starting in state $s$ and performing $k$ actions. For this value function it holds that

$$V(s; k) = R(s, k) + \max_a \sum_{s'} P(s' \mid s, a) V(s'; k - 1),$$

where $P(s' \mid s, a)$ is the probability of transiting to state $s'$ when performing action $a$ in state $s$, and $R(s, k)$ is the immediate reward obtained in state $s$ if $k$ steps are still left. This recursion, which is similar to Eq. (5), is exploited by the value iteration algorithm for solving MDPs. The conditional planning problem with unit observation cost (i.e., $\beta(\mathcal{A}) = |\mathcal{A}|$) could be modeled as a finite-horizon MDP, where states correspond to observed evidence $\mathcal{X}_{\mathcal{A}} = \mathbf{x}_{\mathcal{A}}$, actions correspond to observing variables (or not making any observation) and transition probabilities are given by the probability of observing a particular instantiation of the selected variable. The immediate reward is $R(s, k) = 0$ for $k > 0$, and $R(s, 0)$ is the expected reward (in the value of information problem) of observing assignment $s$ (i.e., $R(P(\mathcal{X}_{\mathcal{V}} \mid s)) - C(s)$). If all observations have unit cost, then for this MDP, it holds that $V(\mathbf{x}_{\mathcal{A}}; k) = J(\mathbf{x}_{\mathcal{A}}; k)$. Unfortunately, in the conditional planning problem, since the state of the MDP is uniquely determined by the observed evidence $\mathcal{X}_{\mathcal{A}} = \mathbf{x}_{\mathcal{A}}$, the state space is *exponentially large*. Hence, existing algorithms for solving MDPs exactly (such as value iteration) cannot be applied to solve large value of information problems. In Section 4.2.2, we develop an efficient dynamic programming algorithm for conditional planning in chain graphical models that avoids this exponential increase in complexity.

### 4.2.2 Dynamic Programming for Optimal Conditional Planning in Chains

We propose a dynamic programming algorithm for obtaining the optimal conditional plan that is similar to the subset algorithm presented in Section 4.1. Again, we utilize the decomposition of rewards described in Section 3. The difference here is that the observation selection and budget allocation now depend on the actual values of the observations. In order to compute the value function $J(\mathbf{x}_{\mathcal{A}}; k)$ for the entire chain, we will compute the value functions $J_{a:b}(\mathbf{x}_{\mathcal{A}}; k)$ for sub-chains $\mathcal{X}_a, \ldots, \mathcal{X}_b$.

The base case of our dynamic programming approach deals with the zero budget setting:

$$J_{a:b}^{flt}(x_a; 0) = \sum_{j=a+1}^{b-1} R_j(\mathcal{X}_j \mid \mathcal{X}_a = x_a),$$

for filtering, and

$$J_{a:b}^{sm}(x_a, x_b; 0) = \sum_{j=a+1}^{b-1} R_j(\mathcal{X}_j \mid \mathcal{X}_a = x_a, \mathcal{X}_b = x_b),$$

for smoothing. The recursion defines $J_{a:b}(x_a; k)$ (or $J_{a:b}(x_a, x_b; k)$ for smoothing), the expected reward for the problem restricted to the sub-chain $\mathcal{X}_a, \ldots, \mathcal{X}_b$ conditioned on the values of $\mathcal{X}_a = x_a$ (and $\mathcal{X}_b = x_b$ for smoothing), and with budget limited by $k$. To compute this quantity, we again iterate through possible split points $j$, such that $a < j < b$. Here we observe a notable difference between the filtering and the smoothing case. For smoothing, we now must consider all possible splits of the budget between the two resulting sub-chains, since an observation at time $j$ might require us to make an additional, earlier observation:

$$J_{a:b}^{sm}(x_a, x_b; k) = \max\Bigg\{ J_{a:b}^{sm}(x_a, x_b; 0), \max_{a < j < b} \Bigg\{ \sum_{x_j} P(\mathcal{X}_j = x_j \mid \mathcal{X}_a = x_a, \mathcal{X}_b = x_b) \Bigg\{$$

$$R_j(\mathcal{X}_j \mid x_j) - C_j(x_j) + \max_{0 \le l \le k - \beta_j(x_j)} \left[ J_{a:j}^{sm}(x_a, x_j; l) + J_{j:b}^{sm}(x_j, x_b; k - l - \beta_j(x_j)) \right] \Bigg\} \Bigg\} \Bigg\}.$$

Looking back in time is not possible in the filtering case, hence the recursion simplifies to

$$J_{a:b}^{flt}(x_a; k) = \max\Bigg\{ J_{a:b}^{flt}(x_a; 0), \max_{a < j < b : \beta_j(x_j) \le k} \Bigg\{ \sum_{x_j} P(\mathcal{X}_j = x_j \mid \mathcal{X}_a = x_a) \Bigg\{$$

$$R_j(\mathcal{X}_j \mid x_j) - C_j(x_j) + J_{a:j}^{flt}(x_a; 0) + J_{j:b}^{flt}(x_j; k - \beta_j(x_j)) \Bigg\} \Bigg\} \Bigg\}.$$

For both $J^{flt}$ and $J^{sm}$, the optimal reward is obtained by $J_{0:n+1}(\emptyset; B) = J(\emptyset; B) = J(\pi^*)$. Algorithm 2 presents a pseudo code implementation for the smoothing version – the filtering case is a straight-forward modification. We call Algorithm 2 the VoIDP *algorithm for conditional planning*. The algorithm will fill the dynamic programming tables using three loops, the inner loop ranging over all assignments $x_a, x_b$, the middle loop ranging over all pairs $(a, b)$ where $a < b$, and the outer loop covers increasing values of $k \le B$. Within the innermost loop, the algorithm again computes a table $sel$ such that $sel(j)$ is the optimal reward achievable by selecting variable $j$ next.

This value is now an expectation over any possible observation that variable $\mathcal{X}_j$ can make. Note that for every possible instantiation $\mathcal{X}_j = x_j$ a different allocation of the remaining budget $k - \beta_j(x_j)$ to the left and right sub-chain ($a : j$ and $j : b$ respectively) can be chosen. The quantity $\sigma(j, x_j)$ tracks this optimal budget allocation.

---

**Input**: Budget $B$, rewards $R_j$, costs $\beta_j$ and penalties $C_j$

**Output**: Optimal conditional plan ($\pi_{a:b}, \sigma_{a:b}$)

**begin**

    **for** $0 \leq a < b \leq n+1, x_a \in \mathrm{dom}\, \mathcal{X}_a, x_b \in \mathrm{dom}\, \mathcal{X}_b$ **do** compute $J_{a:b}^{sm}(x_a, x_b; 0)$;

    **for** $k = 1$ **to** $B$ **do**

        **for** $0 \leq a < b \leq n+1, x_a \in \mathrm{dom}\, \mathcal{X}_a, x_b \in \mathrm{dom}\, \mathcal{X}_b$ **do**

            $sel(-1) := J_{a:b}^{sm}(0)$;

            **for** $a < j < b$ **do**

                $sel(j) := 0$;

                **for** $x_j \in \mathrm{dom}\, \mathcal{X}_j$ **do**

                    **for** $0 \leq l \leq k - \beta_j(x_j)$ **do**

                        $bd(l) := \quad J_{a:j}^{sm}(x_a, x_j; l) + J_{j:b}^{sm}(x_j, x_b; k - l - \beta_j(x_j))$;

                    **end**

                    $sel(j) := sel(j) + P(x_j \mid x_a, x_b) \cdot [R_j(\mathcal{X}_j \mid x_j) - C_j(x_j) + \max_l bd(j)]$;

                    $\sigma(j, x_j) = \mathrm{argmax}_l\, bd(j)$;

                **end**

            **end**

            $J_{a:b}^{sm}(k) = \max_{j \in \{a+1, \ldots, b-1, -1\}} sel(j)$;

            $\pi_{a:b}(x_a, x_b; k) = \mathrm{argmax}_{j \in \{a+1, \ldots, b-1, -1\}} sel(j)$;

            **for** $x_j \in \mathrm{dom}\, X_{\pi_{a:b}(k)}$ **do** $\sigma_{a:b}(x_a, x_b, x_j; k) = \sigma(\pi_{a:b}(k), x_j)$;

        **end**

        **end**

    **end**

**end**

**Algorithm 2**: VOIDP algorithm for computing an optimal conditional plan (for the smoothing setting).

---

**Input**: Budget $k$, observations $\mathcal{X}_a = x_a$, $\mathcal{X}_b = x_b$, $\sigma$, $\pi$

**begin**

    $j := \pi_{a:b}(x_a, x_b; k)$;

    **if** $j \geq 0$ **then**

        Observe $\mathcal{X}_j = x_j$;

        $l := \sigma_{a:b}(x_a, x_b, x_j; k)$;

        Recurse with $k := l$, $\mathcal{X}_a = x_a$ and $\mathcal{X}_j = x_j$ instead of $\mathcal{X}_b = x_b$;

        Recurse with $k := k - l - \beta_j$, $\mathcal{X}_j = x_j$ instead of $\mathcal{X}_a = x_a$, and $\mathcal{X}_b = x_b$;

    **end**

**end**

**Algorithm 3**: Observation selection using conditional planning.

The plan itself is compactly encoded in the quantities $\pi_{a:b}$ and $\sigma_{a:b}$. Hereby, $\pi_{a:b}(x_a, x_b; k)$ determines the next variable to query after observing $\mathcal{X}_a = x_a$ and $\mathcal{X}_b = x_b$, and with remaining budget $k$. $\sigma_{a:b}(x_a, x_b, x_j; k)$ determines the allocation of the budget after the new observation $\mathcal{X}_j = x_j$ has been made. Considering the exponential number of possible sequences of observations, it is remarkable that the optimal plan can be represented using only polynomial space. Algorithm 3 indicates how the computed plan can be executed. The procedure is recursive, requiring the parameters $a := 0$, $x_a := 1$, $b := n + 1$, $x_b := 1$ and $k := B$ for the initial call. In our temperature monitoring example, we could first collect some temperature timeseries as training data, and then learn the chain model from this data. Offline, we would then compute the conditional plan (for the filtering setting), and encode it in the quantities $\pi_{a:b}$ and $\sigma_{a:b}$. We would then deploy the computed plan on the actual sensor node, together with an implementation of Algorithm 3. While computation of the optimal plan (Algorithm 2) is fairly computationally expensive, the execution of the plan (Algorithm 3) is very efficient (selecting the next timestep for observation requires a single lookup in the $\pi_{a:b}$ and $\sigma_{a:b}$ tables) and hence well-suited for deployment on a small, embedded device.

We summarize our analysis in the following Theorem:

**Theorem 2** (**Conditional Planning**). *The algorithm for smoothing presented above computes an optimal conditional plan in* $d^3 \cdot B^2 \cdot (\frac{1}{6}n^3 + \mathcal{O}(n^2))$ *evaluations of local rewards, where $d$ is the maximum domain size of the random variables $X_1, \ldots, X_n$. In the filtering case, the optimal plan can be computed using* $d^3 \cdot B \cdot (\frac{1}{6}n^3 + \mathcal{O}(n^2))$ *evaluations, or, if no budget is used, in* $d^3 \cdot (\frac{1}{6}n^4 + \mathcal{O}(n^3))$ *evaluations.* □

The faster computation for the filtering / no-budget case is obtained by observing that we do not require the third maximum computation, which distributes the budget into the sub-chains.

Also, note that contrary to the algorithm for computing optimal subsets in Section 4.1, Algorithm 2 only requires evaluations of the form $R(\mathcal{X}_j \mid \mathcal{X}_\mathcal{A} = \mathbf{x}_\mathcal{A})$, which can in general be computed $d^2$ times faster than the expectations $R(X_j \mid \mathcal{X}_\mathcal{A})$. Under this consideration, the subset selection algorithm is in general only a factor $d \cdot B$ faster, even though the conditional planning algorithm has more nested loops.

## 4.3 Efficient Algorithms for Trees with Few Leaves

In Sections 4.1 and 4.2 we have presented dynamic programming-based algorithms that can optimize value of information on chain graphical models. In fact, the key observations of Section 3 that local rewards decompose along chains holds not just in chain graphical models, but also in trees.

More formally, a tree graphical model is a joint probability distribution $P(\mathcal{X}_\mathcal{V})$ over a collection of random variables $\mathcal{X}_\mathcal{V}$ if $P(\mathcal{X}_\mathcal{V})$ factors as

$$P(\mathcal{X}_\mathcal{V}) = \frac{1}{Z} \prod_{(i,j) \in \mathcal{E}} \psi_{i,j}(\mathcal{X}_i, \mathcal{X}_j),$$

where $\psi_{i,j}$ is a nonnegative potential function, mapping assignments to $x_i$ and $x_j$ to the nonnegative real numbers, $\mathcal{E} \subseteq \mathcal{V} \times \mathcal{V}$ is a set of edges that form an undirected tree over the index set $\mathcal{V}$, and $Z$ is a normalization constant enforcing a valid probability distribution.

The dynamic programming algorithms presented in the previous sections can be extended to such tree models in a straightforward manner. Instead of identifying optimal subsets and conditional plans for sub-chains, the algorithms would then select optimal subsets and plans for sub-trees of increasing size. Note however that the number of sub-trees can grow exponentially in the number of leaves of the tree: A star on $n$ leaves for example has a number of subtrees that is exponential in $n$. In fact, counting the number of subtrees of an arbitrary tree with $n$ vertices is believed to be intractable (#**P**-complete, Goldberg & Jerrum, 2000). However, for trees that contain only a small (constant) number of leaves, the number of subtrees is polynomial, and the optimal subset and conditional plans can be computed in polynomial time.

## 5. Theoretical Limits

Many problems that can be solved efficiently for discrete chain graphical models can also be efficiently solved for discrete polytrees[3]. Examples include probabilistic inference and the most probable explanation (MPE).

In Section 4.3 however we have seen that the complexity of the dynamic programming algorithms for chains increases dramatically when extended to trees: The complexity increases exponentially in the number of leafs of the tree.

We prove that, perhaps surprisingly, for the problem of optimizing value of information, this exponential increase in complexity cannot be avoided, under reasonable complexity theoretic assumptions. Before making this statement more formal, we briefly review the complexity classes used in our results.

### 5.1 Brief Review of Relevant Computational Complexity Classes

We briefly review the complexity classes used in the following statements by presenting a complete problem for each of the class. For more details see, e.g., the references by Papadimitriou (1995) or Littman, Goldsmith, and Mundhenk (1998). The class $\mathbf{NP}$ contains decision problems which have polynomial-time verifiable proofs. A well-known complete problem is $3SAT$ for which the instances are Boolean formulas $\phi$ in conjunctive normal form containing at most three literals per clause (3CNF form). The complexity class $\#\mathbf{P}$ contains counting problems. A complete problem for the class $\#\mathbf{P}$ is $\#3SAT$ which counts the number of satisfying instances to a 3CNF formula. $\mathbf{PP}$ is a decision version of the class $\#\mathbf{P}$: A complete problem is $MAJSAT$, which decides whether a given 3CNF formula $\phi$ is satisfied by the majority, i.e., by more than half of all its possible assignments. If $A$ and $B$ are Turing machine based complexity classes, then $A^B$ is the complexity class derived by allowing the Turing machines deciding instances of $A$ oracle calls to Turing machines in $B$. We can intuitively think of the problems in class $A^B$ as those that can be solved by a Turing Machine for class $A$, that has a special command which solves any problem in $B$. $\mathbf{PP}$ is similar to $\#\mathbf{P}$ in that $\mathbf{P^{PP}} = \mathbf{P^{\#P}}$, i.e., if we allow a deterministic polynomial time Turing machine to have access to a counting oracle, we cannot solve more complex problems than if we give it access to a majority oracle. Combining these ideas, the class $\mathbf{NP^{PP}}$ is the class of problems that can be solved by nondeterministic polynomial time Turing machines that have access to a majority (or a counting) oracle. A complete problem for $\mathbf{NP^{PP}}$ is $EMAJSAT$ which, given a 3CNF on variables $X_1, \ldots, X_{2n}$, it decides whether there exists an assignment to $X_1, \ldots, X_n$ such that $\phi$ is satisfied for the majority of assignments to $X_{n+1}, \ldots, X_{2n}$. $\mathbf{NP^{PP}}$ has been introduced and found to be a natural class for modeling AI planning problems in the seminal work by Littman et al. (1998). As an example, the MAP assignment problem is $\mathbf{NP^{PP}}$-complete for general graphical models, as shown by Park and Darwiche (2004).

The complexity classes satisfy the following set of inclusions (where the inclusions are assumed, but not known to be strict):

$$\mathbf{P} \subseteq \mathbf{NP} \subseteq \mathbf{PP} \subseteq \mathbf{P^{PP}} = \mathbf{P^{\#P}} \subseteq \mathbf{NP^{PP}}.$$

---

3. Polytrees are Bayesian Networks that form trees if the edge directions are dropped.

## 5.2 Complexity of Computing and Optimizing Value of Information

In order to solve the optimization problems, we will most likely have to evaluate the objective function, i.e., the expected local rewards. Our first result states that, even if we specialize to decision theoretic value of information objective functions as defined in Section 2.1, this problem is intractable even for Naive Bayes models, a special case of discrete polytrees. Naive Bayes models are often used in classification tasks (*c.f.*, Domingos & Pazzani, 1997), where the class variable is predicted from noisy observations (features), that are assumed to be conditionally independent given the class variable. In a sense, Naive Bayes models are the "next simplest" (from the perspective of inference) class of Bayesian networks after chains. Note that Naive Bayes models correspond to the "stars" referred to in Section 4.3, that have a number of subtrees that is exponential in the number of variables.

**Theorem 3 (Hardness of computation for Naive Bayes models).** *The computation of decision theoretic value of information functions is* #**P**-*complete even for Naive Bayes models. It is also hard to approximate to any factor unless* **P** = **NP**. □

We have the immediate corollary that the subset selection problem is **PP**-hard for Naive Bayes models:

**Corollary 4 (Hardness of subset selection for Naive Bayes models).** *The problem of determining, given a Naive Bayes model, constants $c$ and $B$, cost function $\beta$ and a set of decision-theoretic value of information objective functions $R_i$, whether there is a subset of variables $\mathcal{A} \subseteq \mathcal{V}$ such that $L(\mathcal{A}) \geq c$ and $\beta(\mathcal{A}) \leq B$ is* **PP**-*hard.* □

In fact, we can show that subset selection for arbitrary discrete polytrees (that are more general than Naive Bayes models, but inference is still tractable) is even **NP**$^{\mathbf{PP}}$-complete, a complexity class containing problems that are believed to be significantly harder than **NP** or #**P** complete problems. This result provides a complexity theoretic classification of value of information, a core AI problem.

**Theorem 5 (Hardness of subset selection computation for polytrees).** *The problem of determining, given a discrete polytree, constants $c$ and $B$, cost function $\beta$ and a set of decision-theoretic value of information objective functions $R_i$, whether there is a subset of variables $\mathcal{A} \subseteq \mathcal{V}$ such that $L(\mathcal{A}) \geq c$ and $\beta(\mathcal{A}) \leq B$ is* **NP**$^{\mathbf{PP}}$-*complete.* □

For our running example, this implies that the generalized problem of optimally selecting $k$ sensors from a network of correlated sensors is most likely computationally intractable without resorting to heuristics. A corollary extends the hardness of subset selection to the hardness of conditional plans.

**Corollary 6 (Hardness of conditional planning computation for polytrees).** *Computing conditional plans is* **PP**-*hard for Naive Bayes models and* **NP**$^{\mathbf{PP}}$-*hard for discrete polytrees.* □

All proofs of results in this section are stated in the Appendix. They rely on reductions of complete problems in **NP**, #**P** and **NP**$^{\mathbf{PP}}$ involving boolean formulae to problems of computing / optimizing value of information. The reductions are inspired by the works of Littman et al. (1998) and Park and Darwiche (2004), but require the development of novel techniques, such as new reductions of Boolean formulae to Naive Bayes and polytree graphical models associated with appropriate reward functions, ensuring that observation selections lead to feasible assignments to the Boolean formulae.

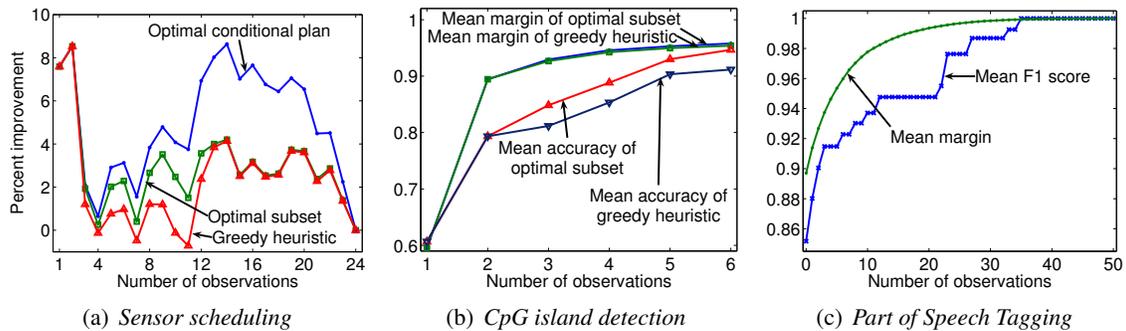

Figure 3: Experimental results. (a) Temperature data: Improvement over the uniform spacing heuristic. (b) CpG island data set: Effect of increasing the number of observations on margin and classification accuracy. (c) Part-of-Speech tagging data set: Effect of increasing the number of observations on margin and F1 score.

## 6. Experiments

In this section, we evaluate our algorithms on several real world data sets. A special focus is on the comparison of the optimal methods with the greedy heuristic and other heuristic methods for selecting observations, and on how the algorithms can be used for interactive structured classification.

### 6.1 Temperature Time Series

The first data set consists of temperature time series collected from a sensor network deployed at Intel Research Berkeley (Deshpande et al., 2004) as described in our running example. Data was continuously collected for 19 days, linear interpolation was used in case of missing samples. The temperature was measured once every 60 minutes, and it was discretized into 10 bins of 2 degrees Kelvin. To avoid overfitting, we used pseudo counts $\alpha = 0.5$ when learning the model. Using parameter sharing, we learned four sets of transition probabilities: from 12 am - 7am, 7 am - 12 pm, 12 pm - 7 pm and 7 pm - 12 am. Combining the data from three adjacent sensors, we got 53 sample time series.

The goal of this task was to select $k$ out of 24 time points during the day, during which sensor readings are most informative. The experiment was designed to compare the performance of the optimal algorithms, the greedy heuristic, and a uniform spacing heuristic, which distributed the $k$ observations uniformly over the day. Figure 3(a) shows the relative improvement of the optimal algorithms and the greedy heuristic over the uniform spacing heuristic. The performance is measured in decrease of expected entropy, with zero observations as the baseline. It can be seen that if $k$ is less than about the half of all possible observations, the optimal algorithms decreased the expected uncertainty by several percent over both heuristics. The improvement gained by the optimal plan over the subset selection algorithms appears to become more drastic if a large number of observations (over half of all possible observations) is allowed. Furthermore, for a large number of observations, the optimal subset and the subset selected by the greedy heuristic were almost identical.

## 6.2 CpG-Island Detection

We then studied the bioinformatics problem of finding CpG islands in DNA sequences. CpG islands are regions in the genome with a high concentration of the cytosine-guanine sequence. These areas are believed to be mainly located around the promoters of genes, which are frequently expressed in the cell. In our experiment, we considered the gene loci HS381K22, AF047825 and AL133174, for which the GenBank annotation listed three, two and one CpG islands each. We ran our algorithm on a 50 base window at the beginning and end of each island, using the transition and emission probabilities from Durbin, Eddy, Krogh, and Mitchison (1999) for our Hidden Markov Model, and we used the sum of margins as reward function.

The goal of this experiment was to locate the beginning and ending of the CpG islands more precisely by asking experts, whether or not certain bases belong to the CpG region or not. Figure 3(b) shows the mean classification accuracy and mean margin scores for an increasing number of observations. The results indicate that, although the expected margin scores are similar for the optimal algorithm and the greedy heuristic, the mean classification performance of the optimal algorithm was still better than the performance of the greedy heuristic. For example, when making 6 observations, the mean classification error obtained by the optimal algorithm is 25% lower than the error obtained by the greedy heuristic.

## 6.3 Part-of-Speech Tagging

In our third experiment, we investigated the structured classification task of part-of-speech (POS) tagging (CoNLL, 2003). Problem instances are sequences of words (sentences), where each word is part of an entity (e.g., "European Union"), and each entity belongs to one of five categories: Location, Miscellaneous, Organization, Person or Other. Imagine an application, where automatic information extraction is guided by an expert: Our algorithms compute an optimal conditional plan for asking the expert, trying to optimize classification performance while requiring as little expert interaction as possible.

We used a conditional random field for the structured classification task, where each node corresponds to a word, and the joint distribution is described by node potentials and edge potentials. The sum of margins was used as reward function. Measure of classification performance was the F1 score, the geometric mean of precision and recall. The goal of this experiment was to analyze how the addition of expert labels increases the classification performance, and how the indirect, decomposing reward function used in our algorithms corresponds to real world classification performance.

Figure 3(c) shows the increase of the mean expected margin and F1 score for an increasing number of observations, summarized over ten 50 word sequences. It can be seen that the classification performance can be effectively enhanced by optimally incorporating expert labels. Requesting only three out of 50 labels increased the mean F1 score from by more than five percent. The following example illustrates this effect: In one scenario both words of an entity, the sportsman 'P. Simmons', were classified incorrectly – 'P.' as *Other* and 'Simmons' as *Miscellaneous*. The first request of the optimal conditional plan was to label 'Simmons'. Upon labeling this word correctly, the word 'P.' was automatically labeled correctly also, resulting in an F1 score of 100 percent.

## 7. Applying Chain Algorithms for More General Graphical Models

In Section 4 we have seen algorithms that can be used to schedule a single sensor, assuming the time series of sensor readings (e.g., temperature) form a Markov chain. This is a very natural assumption for sensor networks (Deshpande et al., 2004). When deploying sensor networks however, multiple sensors need to be scheduled. If the time series for all the sensors were independent, we could use our algorithms to schedule all the sensors independently of each other. However, in practice, the measurements will be correlated across the different sensors – in fact, this dependence is essential to allow generalization of measurements to locations where no sensor has been placed. In the following, we will describe an approach for using our single-sensor scheduling algorithm to coordinate multiple sensors.

More formally, we are interested in monitoring a spatiotemporal phenomenon at a set of locations $\mathcal{S} = \{1, \ldots, m\}$, and time steps $\mathcal{T} = \{1, \ldots, T\}$. With each location–time pair $s, t$, we associate a random variable $\mathcal{X}_{s,t}$ that describes the state of the phenomenon at that location and time. The random vector $\mathcal{X}_{\mathcal{S},\mathcal{T}}$ fully describes the relevant state of the world and the vector $\mathcal{X}_{\mathcal{S},t}$ describes the state at a particular time step $t$. As before, we make the Markov assumption, assuming conditional independence of $\mathcal{X}_{\mathcal{S},t}$ from $\mathcal{X}_{\mathcal{S},t'}$ given $\mathcal{X}_{\mathcal{S},t-1}$ for all $t' < t - 1$.

Similarly as in the single-chain case, we consider reward functions $R_{s,t}$ that are associated with each variable $\mathcal{X}_{s,t}$. Our goal is then to select, for each timestep, a set $\mathcal{A}_t \subseteq \mathcal{S}$ of sensors to activate, in order to maximize the sum of expected rewards. Letting $\mathcal{A}_{1:t} = \mathcal{A}_1 \cup \cdots \cup \mathcal{A}_t$, the expected total reward is then given as

$$\sum_{s,t} R_{s,t}(\mathcal{X}_{s,t} \mid \mathcal{X}_{\mathcal{A}_{1:t}})$$

for the filtering setting (i.e., only observations in the past are taken into account for evaluating the rewards), and

$$\sum_{s,t} R_{s,t}(\mathcal{X}_{s,t} \mid \mathcal{X}_{\mathcal{A}_{1:T}})$$

for the smoothing setting (where all observations are taken into account). The generalization to conditional planning is done as described in Section 2.

Note that in the case of a single sensor ($\ell = 1$), the problem of optimal sensor scheduling can be solved using Algorithm 1. Unfortunately, the optimization problem is wildly intractable even for the case of two sensors, $\ell = 2$:

**Corollary 7 (Hardness of sensor selection for two chains).** *Given a model with two dependent chains, constants $c$ and $B$, a cost function $\beta$ and a set of decision theoretic value of information functions $R_{s,t}$, it is* **NP$^{\text{PP}}$***-complete to determine whether there is a subset $\mathcal{A}_{1:T}$ of variables such that $L(\mathcal{A}_{1:T}) \geq c$ and $\beta(\mathcal{A}_{1:T}) \leq B$.*

In the following, we will develop an approximate algorithm that uses our optimal single-chain algorithms and performs well in practice.

### 7.1 Approximate Sensor Scheduling by Lower Bound Maximization

The reason for the sudden increase in complexity in the case of multiple chains is that the decomposition of rewards along sub-chains (as described in Section 3) does not extend to the case of multiple sensors, since influence can flow across chains. Figure 4 visualizes this problem – there, the distribution for sensor (2) depends on all three observations $S_1^{(1)}$ and $S_4^{(1)}$ from sensor (1) and $S_2^{(2)}$ from sensor (2).

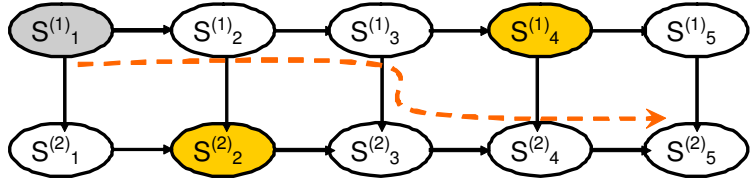

Figure 4: Scheduling multiple correlated sensors in dynamic processes.

We address this complexity issue using an (approximate) extension of the decomposition approach used for single chains. We will focus on the decision-theoretic value of information objective (as described in Section 2.1), but other local reward functions, such as residual entropy, can be used as well.

**Considering only recent observations.** As a first approximation, we only allow a sensor to take into account the most recent observations. Intuitively, this appears to be a reasonable approximation, especially if the potential scheduling times in $\mathcal{T}$ are reasonably far apart. Formally, when evaluating the local rewards at time $t$, we replace the set of observations up to time $t$, $\mathcal{A}_{1:t} \subseteq \mathcal{T}$ by a subset $\mathcal{A}'_{1:t} \subseteq \mathcal{A}_{1:t}$ such that

$$\mathcal{A}'_{1:t} = \big\{ (s, t) \in \mathcal{A}_{1:t} : t \geq t' \text{ for all } (s, t') \in \mathcal{A}_{1:t} \big\},$$

i.e, for each sensor $s$, only the last observation (with largest time index $t$) is kept. We then approximate $R_{s,t}(\mathcal{X}_{s,t} \mid \mathcal{A}_{1:t})$ by $R_{s,t}(\mathcal{X}_{s,t} \mid \mathcal{A}'_{1:t})$. In Figure 4 for example, where $\mathcal{A}_{1:5} = \{(s_1, 1), (s_2, 2), (s_1, 4)\}$, the total expected utility at time $t_5$ would be computed using only observations $\mathcal{A}'_{1:5} = \{(s_2, 2), (s_1, 4)\}$, i.e., using time $t_4$ for sensor one, and time $t_2$ for sensor two, ignoring influence originating from observation $S_1^{(1)}$ and flowing through the chains as indicated by the dashed arrow. The following proposition proves that this approximation is a lower bound to the true value of information:

**Proposition 8 (Monotonicity of value of information).** *The decision-theoretic value of information $R_{s,t}(\mathcal{A})$ of a set $\mathcal{A}$ of sensors is monotonic in $\mathcal{A}$,*

$$R_{s,t}(\mathcal{A}') \leq R_{s,t}(\mathcal{A})$$

*for all $\mathcal{A}' \subseteq \mathcal{A}$.*

Proposition 8 proves that conditioning only on the most recent observations can only decrease our objective function, hence maximizing this approximate objective implies maximizing a lower bound on the true objective.

**A coordinate ascent approach.** We propose the following heuristic for maximizing the lower bound on the expected utility. Instead of jointly optimizing over all schedules (timesteps selected for each sensor), the algorithm will repeatedly iterate over all sensors. For all sensors $s$, it will optimize the selected observations $\mathcal{A}_{1:T}^s$, holding the schedules for all other sensors fixed. This procedure resembles a coordinate ascent approach, where each "coordinate" ranges over all possible schedules for a fixed sensor $s$.

When optimizing for sensor $s$, the algorithm finds a schedule $\mathcal{A}_{1:T}^s$ such that

$$\mathcal{A}_{1:T}^s = \underset{\mathcal{A}_{1:T}}{\operatorname{argmax}} \sum_{s,t} R_{s,t}\left(\mathcal{X}_{s,t} \mid \mathcal{X}_{\mathcal{A}_{1:t}'} \bigcup_{s' \neq s} \mathcal{X}_{\mathcal{A}_{1:t}^{s'}}\right) \text{ such that } \beta(\mathcal{A}_{1:T}^s) \leq B, \tag{6}$$

i.e., that maximizes, over all schedules $\mathcal{A}_{1:T}$, the sum of expected rewards for all time steps and sensors, given the schedules $\mathcal{A}_{1:T}^{s'}$ for all non-selected sensors $s'$.

**Solving the single-chain optimization problem.** In order to solve the maximization problem (6) for the individual sensors, we use the same dynamic programming approach as introduced in Section 4. The recursive case $L_{a:b}^{flt}(k)$ for $k > 0$ is exactly the same. However, the base case is computed as

$$L_{a:b}^{flt}(0) = \sum_{j=a+1}^{b-1} \sum_s R_{s,j}\left(\mathcal{X}_{s,j} \mid \mathcal{X}_a \bigcup_{s' \neq s} \mathcal{X}_{\mathcal{A}_{1:j}^{s'}}\right),$$

i.e., it takes into account the most recent observation for all non-selected sensors $s'$.

Several remarks need to be made about the computation of the base case $L_{a:b}^{flt}(0)$. First of all, in a naive implementation, the computation of the expected utility

$$R_{s,j}\left(\mathcal{X}_{s,j} \mid \mathcal{X}_a \bigcup_{s' \neq s} \mathcal{X}_{\mathcal{A}_{1:j}^{s'}}\right)$$

requires time exponential in the number of chains. This is the case since, in order to compute the reward $R_{s,t}$, for each chain, all possible observations $\mathcal{X}_{\mathcal{A}_{1:t}^s} = \mathbf{x}_{\mathcal{A}_{1:t}^s}$ that could be made need to be taken into account. This computation requires computing the expectation over the joint distribution $P(\mathcal{X}_{\mathcal{A}_{1:t}'})$, which is exponential in size. This increase in complexity can be avoided using a sampling approximation: Hoeffding's inequality can be used to derive polynomial bounds on sample complexity for approximating the value of information up to arbitrarily small additive error $\varepsilon$, similarly as done in the approach of Krause and Guestrin (2005a)[4]. In practice, a small number of samples appears to provide reasonable performance. Secondly, inference itself becomes intractable with an increasing number of sensors. Approximate inference algorithms such as the algorithm proposed by Boyen and Koller (1998) provide a viable way around this problem.

**Analysis.** Since all sensors maximize the same global objective $L(\mathcal{A}_{1:T})$, the coordinated ascent approach is guaranteed to monotonically increase the global objective with every iteration (ignoring possible errors due to sampling or approximate inference). Hence it must converge (to a local optimum) after a finite number of steps. The procedure is formalized in Algorithm 4.

Although we cannot in general provide performance guarantees for the procedure, we are building on an algorithm that provides an optimal schedule for each sensor in isolation, which should benefit from observations provided by the remaining sensors. Also, note that if the sensors are all independent, Algorithm 4 will obtain the optimal solution. Even if the sensors are correlated, the obtained solution will be at least as good as the solution obtained when scheduling all sensors independently of each other. Algorithm 4 will always converge, and always compute a lower bound on

---

4. An absolute error of at most $\varepsilon$ when evaluating each reward $R_{s,t}$ can accumulate to a total error of at most $|\mathcal{T}||\mathcal{S}|\varepsilon$ for all variables and hence to the error of the optimal schedule.

---

**Input**: Budget $B$
**Output**: Selection $\mathcal{A}_1, \ldots, \mathcal{A}_\ell$ of observation times for each sensor
**begin**
    Select $\mathcal{A}_i$, $1 \leq i \leq \ell$ at random;
    **repeat**
        **for** $i = 1$ **to** $\ell$ **do**
            Use Algorithm 1 to select observations $\mathcal{A}_i$ for sensor $i$, but conditioning on current
            sensor scheduling $\mathcal{A}_j$, $j \neq i$, for remaining sensors;
        **end**
        Compute improvement $\delta$ in total expected utility;
    **until** $\delta$ *small enough* ;
**end**

**Algorithm 4**: Multi-Sensor scheduling.

the expected total utility. Considering the intractability of the general problem even for two chains (*c.f.,* , Corollary 7), these properties are reassuring. In our experiments, the coordinated sensor scheduling performed very well, as discussed in Section 7.2.

### 7.2 Proof of Concept Study on Real Deployment

In the work by Singhvi et al. (2005), we presented an approach for optimizing light control in buildings, with the purpose of satisfying building occupants' preferences about lighting conditions, and simultaneously minimizing energy consumption. In our approach, a wireless sensor network is deployed that monitors the building for environmental conditions (such as the sunlight intensity etc.). The sensors feed their measurements to a building controller that actuates the lighting system (lamps, blinds, etc.) accordingly. At every timestep $t \in \mathcal{T}$, the building controller can choose an action that affects the lighting conditions at all locations $\mathcal{S}$ in the building. Utility functions $U_t(a, \mathbf{x}_{\mathcal{S},t})$ are specified that map the chosen actions and the current lighting levels to a utility value. This utility is chosen to capture both users' preferences about light levels, as well as the energy consumption of the lighting system. Details on the utility functions are described in detail by Singhvi et al..

We evaluated our multi-sensor scheduling approach in a real building controller testbed, as described in detail by Singhvi et al.. In our experiments, we used Algorithm 4 to schedule three sensors, allowing each sensor to choose a subset out of ten time steps (in one-hour intervals during daytime). We varied the number of timesteps during which each sensor is activated, and computed the total energy consumption and total user utility (as defined by Singhvi et al.). Figure 5(a) shows the mean user utility and energy savings achieved, for a number of observations varying from no observations to continuous sensing (10 observations in our discretization)[5]. These results imply that using the predictive model and our active sensing strategy, even a very small number of observations achieves results approximately as good as the results achieved by continuous sensing.

Figure 5(b) presents the mean total utility achieved using no observations, one observation or ten observations per sensor each day. It can be seen that even a single observation per sensor increases the total utility close to the level achieved by continuous sensing. Figure 5(c) shows the mean energy consumption required for the same experiment. Here, the single sensor observation strategy comes even closer to the power savings achieved for continuous sensing.

---

5. Note that in Figure 5(a), energy cost and utility are plotted in different units and should not be directly compared.

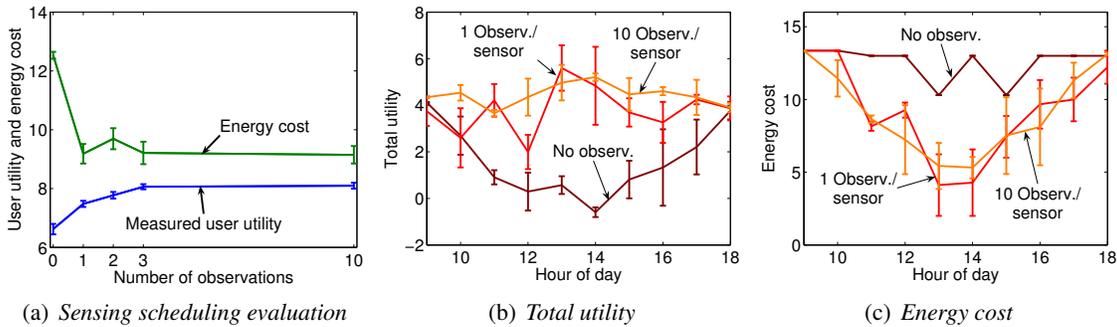

(a) *Sensing scheduling evaluation*  (b) *Total utility*  (c) *Energy cost*

Figure 5: Active sensing results.

Since the sensor network battery lifetime is in general inversely proportional to the amount of power expended for sensing and communication, we conclude that our sensor scheduling strategy promises to lead to drastic increases in sensor network lifetime, deployment permanence and reduced maintenance cost. In our testbed, the network lifetime could be increased by a factor of 3 without significant reduction in user utility and increase in energy cost.

## 8. Related Work

In this section, we review related work in a number of different areas.

### 8.1 Optimal Experimental Design

Optimal experimental design is a general methodology for selecting informative experiments to infer about aspects of the state of the world (such as the parameters of a particular nonlinear function, etc.). There is a large literature about different approaches to experimental design (*c.f.,* Chaloner & Verdinelli, 1995; Krause, Singh, & Guestrin, 2007).

In Bayesian experimental design, a prior distribution over possible states of the world is assumed, and experiments are chosen, e.g., to reduce the uncertainty in the posterior distribution. In its general form, Bayesian experimental design was pioneered by Lindley (1956). The users encode their preferences in a utility function $U(P(\Theta), \theta^\star)$, where the first argument, $P(\Theta)$, is a distribution over states of the world (i.e., the parameters) and the second argument, $\theta^\star$, is the true state of the world. Observations $\mathbf{x}_\mathcal{A}$ are collected, and the change in expected utility under the prior $P(\Theta)$ and posterior $P(\Theta \mid \mathcal{X}_\mathcal{A} = \mathbf{x}_\mathcal{A})$ can be used as a design criterion. In this sense, the value of observation problems considered in this paper can be considered instances of Bayesian experimental design problems. Typically, Bayesian Experimental Design is employed for continuous distributions, often the multivariate normal distribution. By choosing different utility functions, different notions of optimality are defined, including $A$- and $D$- optimality can be developed (Chaloner & Verdinelli, 1995). If we have the posterior covariance matrix $\Sigma_{\theta|A}$, whose maximum eigenvalue is $\lambda_{\max}$, then Bayesian $A$-, $D$-, and $E$- optimality minimizes $\mathrm{tr}\left(\Sigma_{\theta|A}\right)$, $\det\left(\Sigma_{\theta|A}\right)$, and $\lambda_{\max}\left(\Sigma_{\theta|A}\right)$, respectively. In the terminology of Section 2.1, $D$-optimality corresponds to choosing the total entropy, and $A$-optimality corresponds to the (weighted) mean-squared error criteria.

Even for multivariate normal distributions, optimal Bayesian Experimental design is NP-hard (Ko, Lee, & Queyranne, 1995). In some applications of experimental design, the number of experiments to be selected is often large compared to the number of design choices. In these cases, one can find a fractional design (i.e., a non-integral solution defining the proportions by which experiments should be performed), and round the fractional solutions. In the fractional formulation, A-, D-, and E-optimality criteria can be solved exactly using a semi-definite program (Boyd & Vandenberghe, 2004). There are however no known bounds on the integrality gap, i.e., the loss incurred by this rounding process.

The algorithms presented in Section 4.1 can be used to *optimally* solve non-fractional Bayesian Experimental Design problems for chain graphical models, even for continuous distributions, as long as inference in these distributions is tractable (such as normal distributions). This paper hence provides a new class of combinatorial algorithms for an interesting class of Bayesian experimental design problems.

## 8.2 Value of Information in Graphical Models

Decision-theoretic value of information has been frequently used for principled information gathering (*c.f.,* Howard, 1966; Lindley, 1956; Heckerman et al., 1993), and popularized in decision analysis in the context of influence diagrams (Howard & Matheson, 1984). In a sense, value of information problems are special cases of Bayesian experimental design problems, where the prior distribution has a particular structure, typically given by a graphical model as considered in this paper.

Several researchers (Scheffer et al., 2001; van der Gaag & Wessels, 1993; Dittmer & Jensen, 1997; Kapoor et al., 2007) suggested myopic, i.e., greedy approaches for selectively gathering evidence in graphical models, as considered in this paper, which, unlike the algorithms presented in this paper. While these algorithms are applicable to much more general graphical models, they do not have theoretical guarantees. Heckerman et al. (1993) propose a method to compute the maximum expected utility for specific sets of observations. While their work considers more general graphical models than this paper (Naive Bayes models and certain extensions), they provide only large sample guarantees for the *evaluation* of a given sequence of observations, and use a heuristic without guarantees to select such sequences. Bilgic and Getoor (2007) present a branch and bound approach towards exactly optimizing value of information in more complex probabilistic models. In contrast to the algorithms described in this paper however, their approach has running time that is worst-case exponential. Munie and Shoham (2008) present algorithms and hardness results for optimizing a special class of value of information objective functions that are motivated by optimal educational testing problems. Their algorithms apply to a different class of graphical models than chains, and only apply for specific objective functions, rather than general local reward functions as considered in this paper. Radovilsky, Shattah, and Shimony (2006) extended the previous version of our paper (Krause & Guestrin, 2005a) to obtain approximation algorithms with guarantees in the case of noisy observations (i.e., selecting a subset of the emission variables to observe, rather than selecting among the hidden variables as considered in this paper).

## 8.3 Bandit Problems and Exploration / Exploitation

An important class of sequential value of information problems is the class of *Bandit problems*. In the classical $k$-armed bandit problem, as formalized by Robbins (1952), a slot machine is given with $k$ arms. A draw from arm $i$ results in a reward with success probability $p_i$ that is fixed for each arm, but different (and independent) across each arm. When selecting arms to pull, an important problem is to trade off exploration (i.e., estimation of the success probabilities of the arms) and exploitation (i.e., repeatedly pulling the best arm known so far). A celebrated result by Gittins and Jones (1979) shows that for a fixed number of draws, an optimal strategy can be computed in polynomial time, using a dynamic programming based algorithm. While similar in the sense that an optimal sequential strategy can be computed in polynomial time, Gittins algorithm however has different structure from the dynamic programming algorithms presented in this paper.

Note that using the "function optimization" objective function described in Section 2.1, our approach can be used to solve a particular instance of bandit problems, where the arms are not required to be independent, but, in contrary to the classical notion of bandit problems, can not be chosen repeatedly.

### 8.4 Probabilistic Planning

Optimized information gathering has been also extensively studied in the planning community. Bayer-Zubek (2004) for example proposed a heuristic method based on the Markov Decision Process framework. However, her approach makes approximations without theoretical guarantees.

The problem of optimizing decision theoretic value of information can be naturally formalized as a (finite-horizon) Partially Observable Markov Decision Process (POMDP, Smallwood & Sondik, 1973). Hence, in principle, algorithms for planning in POMDPs, such as the anytime algorithm by Pineau, Gordon, and Thrun (2006), can be employed for optimizing value of information. Unfortunately, the state space grows exponentially with the number of variables that are considered in the selection problem. In addition, the complexity of planning in POMDPs grows exponentially in the cardinality of the state space, hence doubly-exponentially in the number of variables for selection. This steep increase in complexity makes application of black-box POMDP solvers infeasible. Recently, Ji, Parr, and Carin (2007) demonstrated the use of POMDP planning on a multi-sensor scheduling problem. While presenting promising empirical results, their approach however uses approximate POMDP planning techniques without theoretical guarantees.

In the robotics literature, Stachniss, Grisetti, and Burgard (2005), Sim and Roy (2005) and Kollar and Roy (2008) have presented approaches to information gathering in the context of Simultaneous Localization and Mapping (SLAM). None of these approaches however provide guarantees about the quality of the obtained solutions. Singh, Krause, Guestrin, Kaiser, and Batalin (2007) present an approximation algorithm with theoretical guarantees for the problem of planning an informative path for environmental monitoring using Gaussian Process models. In contrast to the algorithms presented in this paper, while dealing with more complex probabilistic models and more complex cost functions arising from path planning, their approach requires submodular objective functions (a property that does not hold for value of information as we show in Proposition 9).

### 8.5 Sensor Selection and Scheduling

In the context of wireless sensor networks, where sensor nodes have limited battery and can hence only enable a small number of measurements, optimizing the value of information from the selected sensors plays a key role. The problem of deciding when to selectively turn on sensors in order to conserve power was first discussed by Slijepcevic and Potkonjak (2001) and Zhao, Shin, and Reich (2002). Typically, it is assumed that sensors are associated with a fixed sensing region, and a spatial domain needs to be covered by the regions associated with the selected sensors. Abrams, Goel, and Plotkin (2004) present an efficient approximation algorithm with theoretical guarantees for this problem. Deshpande, Khuller, Malekian, and Toossi (2008) present an approach for this problem based on semidefinite programming (SDP), handling more general constraints and providing tighter approximations. The approaches described above do not apply to the problem of optimizing sensor schedules for more complex utility functions such as, e.g., the increase in prediction accuracy and other objectives considered in this paper. To address these shortcomings, Koushanfary, Taft, and Potkonjak (2006) developed an approach for sensor scheduling that guarantees a specified prediction accuracy based on a regression model. However, their approach relies on the solution of a Mixed Integer Program, which is intractable in general. Zhao et al. (2002) proposed heuristics for selectively querying nodes in a sensor network in order to reduce the entropy of the prediction. Unlike the algorithms presented in this paper, their approaches do not have any performance guarantees.

## 8.6 Relationship to Machine Learning

Decision Trees (Quinlan, 1986) popularized the value of information as a criterion for creating conditional plans. Unfortunately, there are no guarantees on the performance of this greedy method.

The subset selection problem as an instance of feature selection is a central issue in machine learning, with a vast amount of literature (see Molina, Belanche, & Nebot, 2002 for a survey). However, we are not aware of any work providing similarly strong performance guarantees than the algorithms considered in this paper.

The problem of choosing observations also has a strong connection to the field of active learning (*c.f.,* Cohn, Gharamani, & Jordan, 1996; Tong & Koller, 2001) in which the learning system designs experiments based on its observations. While sample complexity bounds have been derived for some active learning problems (*c.f.,* Dasgupta, 2005; Balcan, Beygelzimer, & Langford, 2006), we are not aware of any active learning algorithms that perform provably optimal (even for restricted classes of problem instances).

## 8.7 Previous Work by the Authors

A previous version of this paper appeared in the work by Krause and Guestrin (2005b). Some of the contents of Section 7 appeared as part of the work by Singhvi et al. (2005). The present version is much extended, with new algorithmic and hardness results and more detailed discussions.

In light of the negative results presented in Section 5, we cannot expect to be able to optimize value of information in more complex models than chains. However, instead of attempting to solve for the optimal solution, one might wonder whether it is possible to obtain good approximations. The authors showed (Krause & Guestrin, 2005a; Krause et al., 2007; Krause, Leskovec, Guestrin, VanBriesen, & Faloutsos, 2008) that a large number of practical objective functions satisfy an intuitive diminishing returns property: Adding a new observation helps more if we have few observations so far, and less if we have already made many observations. This intuition can be formalized using the combinatorial concept called *submodularity*. A fundamental result by Nemhauser et al. proves that when optimizing a submodular utility function, the myopic greedy algorithm in fact provides a near-optimal solution, that is within a constant factor of $(1-1/e) \approx 63\%$ of optimal. Unfortunately, decision theoretic value of information does not satisfy submodularity.

**Proposition 9** (**Non-submodularity of value of information**). *Decision-theoretic value of information is not submodular, even in Naive Bayes models.*

Intuitively, value of information can be non-submodular, if we need to make several observations in order to "convince" ourselves that we need to change our action.

## 9. Conclusions

We have described novel efficient algorithms for optimal subset selection and conditional plan computation in chain graphical models (and trees with few leaves), including HMMs. Our empirical evaluation indicates that these algorithms can improve upon commonly used heuristics for decreasing expected uncertainty. Our algorithms can also effectively enhance performance in interactive structured classification tasks.

Unfortunately, the optimization problems become wildly intractable for even a slight generalization of chains. We presented surprising theoretical limits, which indicate that even the class of decision theoretic value of information functions (as widely used, e.g., in influence diagrams and POMDPs) cannot be efficiently computed even in Naive Bayes models. We also identified optimization of value of information as a new class of problems that are intractable (**NP**$^{\mathbf{PP}}$-complete) for polytrees.

Our hardness results, along with other recent results for polytree graphical models, the **NP**-completeness of maximum a posteriori assignment (Park & Darwiche, 2004) and **NP**-hardness of inference in conditional linear Gaussian models (Lerner & Parr, 2001), suggest the possibility of developing a generalized complexity characterization of problems that are hard in polytree graphical models.

In light of these theoretical limits for computing optimal solutions, it is a natural question to ask whether approximation algorithms with non-trivial performance guarantees can be found. Recent results by Krause and Guestrin (2005a), Radovilsky et al. (2006) and Krause et al. (2007) show that this is the case for interesting classes of value of information problems.

## Acknowledgments

We would like to thank Ben Taskar for providing the part-of-speech tagging model, and Reuters for making their news archive available. We would also like to thank Brigham Anderson and Andrew Moore for helpful comments and discussions. This work was partially supported by NSF Grants No. CNS-0509383, CNS-0625518, ARO MURI W911NF0710287 and a gift from Intel. Carlos Guestrin was partly supported by an Alfred P. Sloan Fellowship, an IBM Faculty Fellowship and an ONR Young Investigator Award N00014-08-1-0752 (2008-2011). Andreas Krause was partially supported by a Microsoft Research Graduate Fellowship.

## Appendix A

**Proof of Theorem 3.** Membership in **#P** for arbitrary discrete polytrees is straightforward since inference in such models is in **P**. Let $\phi$ be an instance of $\#3SAT$, where we have to count the number of assignments to $X_1, \ldots, X_n$ satisfying $\phi$. Let $C = \{C_1, \ldots, C_m\}$ be the set of clauses. Now create a Bayesian network with $2n + 1$ variables, $\mathcal{X}_1, \ldots, \mathcal{X}_n, \mathcal{U}_1, \ldots, \mathcal{U}_n$ and $\mathcal{Y}$, where the $\mathcal{X}_i$ are conditionally independent given $\mathcal{Y}$. Let $\mathcal{Y}$ be uniformly distributed over the values $\{-n, -(n-1), \ldots, -1, 1, \ldots, m-1, m\}$, and each $\mathcal{U}_i$ have Bernoulli prior with $p = 0.5$. Let the observed variables $\mathcal{X}_i$ have CPTs defined the following way:

$$\mathcal{X}_i \mid [\mathcal{Y} = +j, \mathcal{U}_i = u] \sim \begin{cases} 1, & \text{if } X_i = u \text{ satisfies clause } C_j; \\ 0, & \text{otherwise.} \end{cases}$$

$$\mathcal{X}_i \mid [\mathcal{Y} = -j, \mathcal{U}_i = u] \sim \begin{cases} 0, & \text{if } i = j; \\ u, & \text{otherwise.} \end{cases}$$

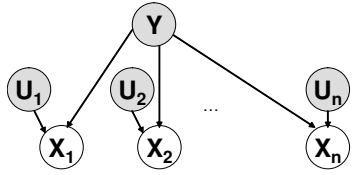

Figure 6: Graphical model used in the proof of Theorem 3.

In this model, which is presented in Figure 6, it holds that $\mathcal{X}_1 = \mathcal{X}_2 = \cdots = \mathcal{X}_n = 1$ iff $\mathcal{U}_1, \ldots, \mathcal{U}_n$ encode a satisfying assignment of $\phi$, and $\mathcal{Y} > 0$. Hence, if we observe $\mathcal{X}_1 = \mathcal{X}_2 = \cdots = \mathcal{X}_n = 1$, we know that $\mathcal{Y} > 0$ with certainty. Furthermore, if at least one $\mathcal{X}_i = 0$, we know that $P(\mathcal{Y} > 0 \mid \mathcal{X} = \mathbf{x}) < 1$. Let all nodes have zero reward, except for $\mathcal{Y}$, which is assigned a reward function with the following properties (we will show below how we can model such a local reward function using the decision-theoretic value of information):

$$R(\mathcal{Y} \mid \mathcal{X}_{\mathcal{A}} = \mathbf{x}_{\mathcal{A}}) = \begin{cases} \frac{(n+m)2^n}{m}, & \text{if } P(\mathcal{Y} > 0 \mid \mathcal{X}_{\mathcal{A}} = \mathbf{x}_{\mathcal{A}}) = 1; \\ 0, & \text{otherwise.} \end{cases}$$

By the above argument, the expected reward

$$R(\mathcal{Y} \mid \mathcal{X}_1, \ldots, \mathcal{X}_n) = \sum_{\mathbf{u}, y, \mathbf{x}} P(\mathcal{Y} = y) P(\mathcal{U} = \mathbf{u}) P(\mathbf{x} \mid \mathbf{u}) R(\mathcal{Y} \mid \mathcal{X} = \mathbf{x})$$

$$= \sum_{\mathbf{u} \text{ sat } \phi} P(\mathcal{Y} > 0) P(\mathbf{u}) \frac{(n+m)2^n}{m} = \sum_{u \text{ sat } \phi} 1$$

is exactly the number of satisfying assignments to $\phi$. Note that the model defined above is not yet a Naive Bayes model. However, it can easily be turned into one by marginalizing out $\mathcal{U}$.

We will now show how we can realize a reward function with the above properties in the maximum expected utility sense. Let $D = \{d_1, d_2\}$ be a set of two decisions. Define a utility function with the property:

$$u(y, d) = \begin{cases} \frac{(n+m)2^n}{m}, & \text{if } d = d_1 \text{ and } y > 0; \\ \frac{(n+m)2^{n+1}}{n}, & \text{if } d = d_1 \text{ and } y < 0; \\ 0, & \text{otherwise.} \end{cases}$$

The reward $R(\mathcal{Y} \mid \mathcal{X}_{\mathcal{A}})$ is then given as the decision-theoretic value of information:

$$R(\mathcal{Y} \mid \mathcal{X}_{\mathcal{A}}) = \sum_{\mathbf{x}_{\mathcal{A}}} P(\mathbf{x}_{\mathcal{A}}) \max_d \sum_y P(y \mid \mathbf{x}_{\mathcal{A}}) u(y, d).$$

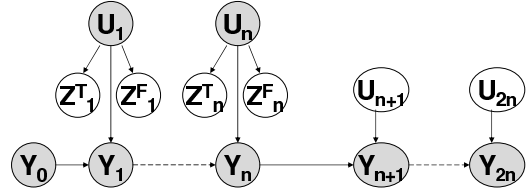

Figure 7: Graphical model used in proof of Theorem 5.

The utility function $u$ is based on the following consideration. Upon observing a particular instantiation of the variables $\mathcal{X}_1, \ldots, \mathcal{X}_n$ we make a decision $d$ about variable $\mathcal{Y}$. Our goal is to achieve that the number of times action $d_1$ is chosen exactly corresponds to the number of satisfying assignments to $\phi$. This is accomplished in the following way. If all $\mathcal{X}_i$ are 1, then we know that the $\mathcal{U}_i$ had encoded a satisfying assignment, and $\mathcal{Y} > 0$ with probability 1. In this case, action $d_1$ is chosen. Now we need to make sure that whenever at least one $\mathcal{X}_i = 0$ (which indicates either that $\mathcal{Y} < 0$ or $\mathcal{U}$ is not a satisfying assignment) decision $d_2$ is chosen. Now, if at least one $\mathcal{X}_i = 0$, then either $\mathcal{Y} = j > 0$ and clause $j$ was not satisfied, or $\mathcal{Y} < 0$. The utilities are designed such that unless $P(\mathcal{Y} > 0 \mid \mathcal{X}_\mathcal{A} = \mathbf{x}_\mathcal{A}) \geq 1 - \frac{n2^{-n}}{2m}$, the action $d_2$ gives the higher expected reward of 0. Hereby, $\frac{n2^{-n}}{2m}$ is a lower bound on the probability of "misclassification" $P(\mathcal{Y} < 0 \mid \mathcal{X}_\mathcal{A} = \mathbf{x}_\mathcal{A})$.

Note that the above construction immediately proves the hardness of approximation: Suppose there were a polynomial time algorithm which computes an approximation $\hat{R}$ that is within any factor $\alpha > 1$ (which can depend on the problem instance) of $R = R(\mathcal{Y} \mid \mathcal{X}_1, \ldots, \mathcal{X}_n)$. Then $\hat{R} > 0$ implies that $R > 0$, and $\hat{R} = 0$ implies that $R = 0$. Hence, the approximation $\hat{R}$ can be used to decide whether $\phi$ is satisfiable or not, implying that $\mathbf{P} = \mathbf{NP}$. □

**Proof of Corollary 4.** Let $\phi$ be a 3CNF formula. We convert it into a Naive Bayes model over variables $\mathcal{X}_1, \ldots, \mathcal{X}_n$ and $\mathcal{Y}$ as in the construction of Theorem 3. The function $L(\mathcal{V})$ where $\mathcal{V} = \{1, \ldots, n\}$ is the set of all variables $\mathcal{X}_i$ counts the number of satisfying assignments to $\phi$. Note that the function $L(\mathcal{A})$ for $\mathcal{A} \subseteq \mathcal{V} = \{1, \ldots, n\}$ is monotonic, i.e., $L(\mathcal{A}) \leq L(\mathcal{V})$ for all $\mathcal{A} \subseteq \mathcal{V}$, as shown in Proposition 8. Hence the majority of assignments satisfies $\phi$ if and only if $L(\mathcal{V}) > 2^{n-1}$. □

**Proof of Theorem 5.** Membership follows from the fact that inference in polytrees is in $\mathbf{P}$ for discrete polytrees: A nondeterministic Turing machine with $\#\mathbf{P}$ oracle can first "guess" the selection of variables, then compute the value of information using Theorem 3 (since such computation is $\#\mathbf{P}$-complete for arbitrary discrete polytrees), and compare against constant $c$.

To show hardness, let $\phi$ be an instance of $EMAJSAT$, where we have to find an instantiation of $X_1, \ldots, X_n$ such that $\phi(X_1, \ldots, X_{2n})$ is true for the majority of assignments to $X_{n+1}, \ldots, X_{2n}$. Let $C = \{C_1, \ldots, C_m\}$ be the set of 3CNF clauses. Create the Bayesian network shown in Figure 7, with nodes $\mathcal{U}_i$, each having a uniform Bernoulli prior. Add bivariate variables $\mathcal{Y}_i = (sel_i, par_i)$, $0 \leq i \leq 2n$, where $sel_i$ takes values in $\{0, \ldots, m\}$ and $par_i$ is a parity bit. The CPTs for $\mathcal{Y}_i$ are defined as: $sel_0$ uniformly varies over $\{1, \dots, m\}$, $par_0 = 0$, and for $\mathcal{Y}_1, \dots, \mathcal{Y}_{2n}$:

$$sel_i \mid [sel_{i-1} = j, \mathcal{U}_i = u_i] \sim \left\{ \begin{array}{ll} 0, & \text{if } j = 0, \text{ or } u_i \text{ satisfies } C_j; \\ j, & \text{otherwise;} \end{array} \right.$$

$$par_i \mid [par_{i-1} = b_{i-1}, \mathcal{U}_i] \sim b_{i-1} \oplus \mathcal{U}_i,$$

where $\oplus$ denotes the parity (XOR) operator. We now add variables $\mathcal{Z}_i^T$ and $\mathcal{Z}_i^F$ for $1 \leq i \leq n$ and let

$$\mathcal{Z}_i^T \mid [\mathcal{U}_i = u_i] \sim \left\{ \begin{array}{ll} \text{Uniform}(\{0, 1\}), & \text{if } u_i = 1; \\ 0, & \text{otherwise;} \end{array} \right.$$

where Uniform denotes the uniform distribution. Similarly, let

$$\mathcal{Z}_i^F \mid [\mathcal{U}_i = u_i] \sim \left\{ \begin{array}{ll} \text{Uniform}(\{0, 1\}), & \text{if } u_i = 0; \\ 0, & \text{otherwise.} \end{array} \right.$$

Intuitively, $\mathcal{Z}_i^T = 1$ guarantees us that $\mathcal{U}_i = 1$, whereas $\mathcal{Z}_i^T = 0$ leaves us uncertain about $\mathcal{U}_i$. The case of $\mathcal{Z}_i^F$ is symmetric.

We use the subset selection algorithm to choose the $\mathcal{Z}_i$s that encode the solution to $EMAJSAT$. If $\mathcal{Z}_i^T$ is chosen, it will indicate that $X_i$ should set to true, similarly $\mathcal{Z}_i^F$ indicates a false assignment to $X_i$. The parity function is going to be used to ensure that exactly one of $\{\mathcal{Z}_i^T, \mathcal{Z}_i^F\}$ is observed for each $i$.

We first assign penalties $\infty$ to all nodes except $\mathcal{Z}_i^T, \mathcal{Z}_i^F$ for $1 \leq i \leq n$, and $\mathcal{U}_j$ for $n + 1 \leq j \leq 2n$, which are assigned zero penalty. Let all nodes have zero reward, except for $\mathcal{Y}_{2n}$, which is assigned the following reward:

$$R(\mathcal{Y}_{2n} \mid \mathcal{X}_\mathcal{A} = \mathbf{x}_\mathcal{A}) = \left\{ \begin{array}{ll} 4^n, & \text{if } P(sel_{2n} = 0 \mid \mathcal{X}_\mathcal{A} = \mathbf{x}_\mathcal{A}) = 1 \text{ and} \\ & [P(par_{2n} = 1 \mid \mathcal{X}_\mathcal{A} = \mathbf{x}_\mathcal{A}) = 1 \text{ or } P(par_{2n} = 0 \mid \mathcal{X}_\mathcal{A} = \mathbf{x}_\mathcal{A}) = 1]; \\ 0, & \text{otherwise.} \end{array} \right.$$

Note that $sel_{2n} = 0$ with probability 1 iff $\mathcal{U}_1, \dots, \mathcal{U}_{2n}$ encode a satisfying assignment of $\phi$. Furthermore, we get positive reward only if we are both certain that $sel_{2n} = 0$, i.e., the chosen observation set must contain a proof that $\phi$ is satisfied, and we are certain about $par_{2n}$. The parity certainty will only occur if we are certain about the assignment $\mathcal{U}_1, \dots, \mathcal{U}_{2n}$. It is only possible to infer the value of each $\mathcal{U}_i$ with certainty by observing one of $\mathcal{U}_i, \mathcal{Z}_i^T$ or $\mathcal{Z}_i^F$. Since, for $i = 1, \dots, n$, the cost of observing $\mathcal{U}_i$ is $\infty$, to receive any reward we must observe at least one of $\mathcal{Z}_i^T$ or $\mathcal{Z}_i^F$. Assume that we compute the optimal subset $\hat{\mathcal{O}}$ for budget $2n$, then we can only receive positive reward by observing exactly one of $\mathcal{Z}_i^T$ or $\mathcal{Z}_i^F$.

We interpret the selection of $\mathcal{Z}_i^T, \mathcal{Z}_i^F$ as an assignment to the first $n$ variables of $EMAJSAT$. Let $\hat{R} = R(\mathcal{Y}_{2n} \mid \hat{\mathcal{O}})$. We claim that $\phi \in EMAJSAT$ if and only if $\hat{R} > 0.5$. First let $\phi \in EMAJSAT$, with assignment $x_1, \dots, x_n$ to the first $n$ variables. Now add $\mathcal{U}_{n+1}, \dots, \mathcal{U}_{2n}$ to $\mathcal{O}$ and add $\mathcal{Z}_i^T$ to $\mathcal{O}$ iff $x_i = 1$ and $\mathcal{Z}_i^F$ to $\mathcal{O}$ iff $x_i = 0$. This selection guarantees $\hat{R} > 0.5$.

Now assume $\hat{R} > 0.5$. We call an assignment to $\mathcal{U}_1, \dots, \mathcal{U}_{2n}$ *consistent* if for any $1 \leq i \leq n$, if $\mathcal{Z}_i^T \in \hat{\mathcal{O}}$, then $\mathcal{U}_i = 1$ and if $\mathcal{Z}_i^F \in \hat{\mathcal{O}}$ then $\mathcal{U}_i = 0$. For any consistent assignment, the chance that the observations $\mathcal{Z}_i$ prove the consistency is $2^{-n}$. Hence $\hat{R} > 0.5$ implies that the majority of all provably consistent assignments satisfy $\phi$ and hence $\phi \in EMAJSAT$. This proves that subset selection is $\mathbf{NP^{PP}}$ complete.

Note that we can realize the local reward function $R$ in the sense of maximum expected utility similarly as described in the Proof of Theorem 3. $\qquad \square$

**Proof of Corollary 6.** The constructions in the proof of Theorem 4 and Theorem 5 also prove that computing conditional plans is **PP**-hard and **NP<sup>PP</sup>**-hard respectively, since, in these instances, any plan with positive reward must observe variables corresponding to valid instantiations (i.e., all $\mathcal{X}_1, \ldots, \mathcal{X}_n$ in Corollary 4, and all $\mathcal{U}_{n+1}, \ldots, \mathcal{U}_{2n}$ and one each of the $\mathcal{Z}_1, \ldots, \mathcal{Z}_n$ to satisfy the parity condition in Theorem 5). In these cases, the order of selection is irrelevant, and, hence, the conditional plan effectively performs subset selection. □

**Proof of Corollary 7.** The proof follows from the observation that polytree construction from the proof of Theorem 5 can be arranged into two dependent chains. For this transformation, we revert the arc between $\mathcal{Z}_i^T$ and $\mathcal{U}_i$ by applying Bayes' rule. To make sure there are the same number of nodes for each sensor in each timeslice, we triple the variables $\mathcal{Y}_i$, calling the copies $\mathcal{Y}_i'$ and $\mathcal{Y}_i''$. The conditional probability tables are given as equality constraints, $\mathcal{Y}_i' = \mathcal{Y}_i$ and $\mathcal{Y}_i'' = \mathcal{Y}_i'$. After this transformation, the variables associated with timesteps $3i - 2$ (for $i \geq 1$) are given by the sets $\{\mathcal{Y}_{i-1}'', \mathcal{Z}_i^T\}$. timesteps $3i - 1$ are associated with the sets $\{\mathcal{U}_i, \mathcal{Y}_i\}$, and timesteps $3i$ are associated with $\{\mathcal{Z}_i^F, \mathcal{Y}_i'\}$. □

**Proof of Proposition 8.** This bound follows from the fact that maximization over $\mathbf{a}$ is convex, and an application of Jensen's inequality. Using an induction argument, we simply need to show that $L(\mathcal{A}) \geq L(\emptyset)$.

$$
\begin{aligned}
L(\mathcal{A}) &= \sum_{\mathbf{x}_\mathcal{A}} P(\mathcal{X}_\mathcal{A} = \mathbf{x}_\mathcal{A}) \left( \sum_{t \in \mathcal{V}} \max_a EU(a, t, \mathbf{x} \mid \mathcal{X}_{\mathcal{A}_{1:t}} = \mathbf{x}_{\mathcal{A}_{1:t}}) \right) \\
&\geq \sum_{t \in \mathcal{V}} \max_a \left( \sum_{\mathbf{x}_\mathcal{A}} P(\mathcal{X}_\mathcal{A} = \mathbf{x}_\mathcal{A}) EU(a, t, \mathbf{x} \mid \mathcal{X}_{\mathcal{A}_{1:t}} = \mathbf{x}_{\mathcal{A}_{1:t}}) \right) \\
&= \sum_{t \in \mathcal{V}} \max_a EU(\mathbf{a}, t, \mathbf{x}) = L(\emptyset)
\end{aligned}
$$

where

$$
EU(a, t, \mathbf{x} \mid \mathcal{X}_{\mathcal{A}_{1:t}} = \mathbf{x}_{\mathcal{A}_{1:t}}) = \sum_{x_t} P(x_t \mid \mathcal{X}_{\mathcal{A}_{1:t}} = \mathbf{x}_{\mathcal{A}_{1:t}}) U_t(a, x_t)
$$

is the expected utility of action $a$ at time $t$ after observing $\mathcal{X}_{\mathcal{A}_{1:t}} = \mathbf{x}_{\mathcal{A}_{1:t}}$. □

**Proof of Proposition 9.** Consider the following binary classification problem with assymetric cost. We have one Bernoulli random variable $\mathcal{Y}$ (the class label) with $P(\mathcal{Y} = 1) = 0.5$ and $P(\mathcal{Y} = -1) = 0.5$. We also have two noisy observations $\mathcal{X}_1, \mathcal{X}_2$, which are conditionally independent given $\mathcal{Y}$. Let $P(\mathcal{X}_i = \mathcal{Y}) = 3/4$ (i.e., the observations agree with the class label with probability 3/4, and disagree with probability 1/4. We have three actions, $a_1$ (classifying $\mathcal{Y}$ as 1), $a_{-1}$ (classifying $\mathcal{Y}$ as -1) and $a_0$ (not assigning any label). We define our utility functon $U$ such that we gain utility 1 if we assign the label correctly ($U(a_1, 1) = U(a_{-1}, -1) = 1$), $-3$ is we misassign the label ($U(a_1, -1) = U(a_{-1}, 1) = -3$), and 0 if we choose $a_0$, i.e., not assign any label. Now, we can verify that $L(\emptyset) = L(\{\mathcal{X}_1\}) = L(\{\mathcal{X}_2\}) = 0$, but $L(\{\mathcal{X}_1, \mathcal{X}_2\}) = \left(\frac{3}{4}\right)^2 - 3\left(\frac{1}{4}\right)^2 = \frac{6}{16} > 0$. Hence, adding $\mathcal{X}_2$ to $\mathcal{X}_1$ increases the utility more than adding $\mathcal{X}_2$ to the empty set, contradicting submodularity. □